\documentclass[fullname]{clv3}

\usepackage{xcolor}

\usepackage{amsmath}

\usepackage{picins}
\usepackage{tikz}

\usepackage{enumitem}

\usepackage{tikz-qtree}

\usepackage{mathtools}

\mathtoolsset{showonlyrefs,showmanualtags}

\definecolor{azure}{rgb}{0.0, 0.5, 1.0}

\newcommand{\nts}{\mathcal{L}}
\newcommand{\vocab}{\mathcal{T}}
\newcommand{\vars}{\mathcal{V}}
\newcommand{\tcol}{\mathrm{ToCol}}
\newcommand{\trow}{\mathrm{ToRow}}
\newcommand{\fcol}{\mathrm{FromCol}}
\newcommand{\urow}{\mathrm{UnmarkRow}}
\newcommand{\ucol}{\mathrm{UnmarkCol}}
\newcommand{\frow}{\mathrm{FromRow}}
\newcommand{\remove}{\mathrm{remove}}
\newcommand{\insertt}{\mathrm{insert}}
\newcommand{\merge}{m}

\newcommand{\rulessym}{\mathcal{R}}

\newcommand{\ijleq}{<}

\newcommand{\citet}{\namecite}

\def\condname{Condition}

\newtheorem{thm}{Theorem}[section]
\newtheorem{cond}[thm]{Condition}

\newcommand{\config}{\mathrm{config}}
\newcommand{\configone}{\mathrm{config}_1}
\newcommand{\configtwo}{\mathrm{config}_2}
\newcommand{\configthree}{\mathrm{config}_3}

\newcommand{\configsym}{c}

\newcommand{\startsym}{S}

\def\abx{\tikz[x=0.75em,y=0.75em,baseline=0]\fill[even odd rule]
(0,0)--(.5,0)--(.5,.45)--(.75,.45)--(.75,.25)--(1,.5)--(.75,.75)--
(.75,.55)--(.5,.55)--(.5,1)--(0,1)--cycle(.1,.1)rectangle(.4,.9);}
\def\aby{\tikz[x=0.75em,y=0.75em,baseline=0]\fill[even odd rule]
(0,0)--(.5,0)--(.5,.5)--(.75,.25)--(.75,.45)--(1,.45)--(1,.55)--(.75,.55)--
(.75,.75)--(.5,.5)--(.5,1)--(0,1)--cycle (.1,.1)rectangle(.4,.9);}
\def\abt{\tikz[x=0.75em,y=0.75em,baseline=0]\fill[even odd rule]
(0,0)--(.5,0)--(.5,.5)--(.5,.5)--(.5,1)--(0,1)--cycle (.1,.1)
rectangle(.4,.9);}
\def\abz{\rotatebox{90}{\abx}}
\def\abw{\rotatebox{90}{\aby}}
\def\abu{\rotatebox{90}{\abt}}

\renewcommand{\tcol}{\aby}
\renewcommand{\frow}{\abz}
\renewcommand{\trow}{\abw}
\renewcommand{\fcol}{\abx}
\renewcommand{\urow}{\abu}
\renewcommand{\ucol}{\abt}

\title{Parsing Linear Context-Free Rewriting Systems with Fast Matrix Multiplication}

\runningtitle{Parsing LCFRS with Fast Matrix Multiplication}

\runningauthor{Cohen and Gildea}

\begin{document}

\author{Shay B.~Cohen\thanks{ School of Informatics, University of Edinburgh, Edinburgh, EH8 9AB, United Kingdom. E-mail: \texttt{scohen@inf.ed.ac.uk}.}}
\affil{University of Edinburgh}

\author{Daniel Gildea\thanks{ Department of Computer Science, University of Rochester, Rochester, NY 14627, United States. E-mail: \texttt{gildea@cs.rochester.edu}.}}
\affil{University of Rochester}
\historydates{Submission received:            26th October, 2016;
Revised version received:               19th January, 2016;
Accepted for publication:               15th February, 2016.}
\maketitle

\begin{abstract}
We describe a recognition algorithm for a subset of binary linear context-free rewriting systems (LCFRS) with running time $O(n^{\omega d})$ where $M(m) = O(m^{\omega})$ is the running
time for $m \times m$ matrix multiplication and $d$ is the ``contact rank'' of the LCFRS -- the maximal number of combination and non-combination points that appear in
the grammar rules. We also show that this algorithm can be used as a subroutine to get a recognition algorithm for general binary LCFRS with running time $O(n^{\omega d + 1})$.
The currently best known $\omega$ is smaller than 2.38. 
Our result provides another proof for the best known result for parsing mildly context sensitive formalisms such as combinatory categorial grammars, head grammars, linear indexed grammars,
and tree-adjoining grammars, which can be parsed in time $O(n^{4.76})$. 
It also shows that inversion transduction grammars can be parsed in time $O(n^{5.76})$. In addition, binary LCFRS subsumes many other formalisms and types of grammars, for some of
which we also improve the asymptotic complexity of parsing.
\end{abstract}

\section{Introduction}
\label{section:introduction}

The problem of {\bf grammar recognition} is a decision problem  
of determining whether a string belongs to a language induced by a grammar.
For context-free grammars, recognition can be done using parsing algorithms such as the CKY algorithm \cite{kasami-65,younger-67,cocke-70} or the Earley algorithm \cite{earley1970efficient}.
The asymptotic complexity of these chart parsing algorithms is cubic in the length of the sentence.

In a major breakthrough, \citet{valiant75} showed that 
context-free grammar recognition is no more complex than
Boolean matrix multiplication for a matrix
of size $m \times m$ where $m$ is linear in the length of the sentence, $n$. With current state-of-the-art results in matrix multiplication, this means
that CFG recognition can be done with an asymptotic complexity of $O(n^{2.38})$.

In this paper, we show that the problem of linear context-free rewriting system recognition can also be reduced to Boolean matrix multiplication.
Current chart parsing algorithms for binary LCFRS have an asymptotic complexity of $O(n^{3f})$, where $f$ is the maximal fan-out of the grammar.\footnote{Without placing a bound on $f$, the
problem of recognition of LCFRS languages is NP-hard \cite{satta1992recognition}.}
Our algorithm takes time $O(n^{\omega d})$, for a constant
$d$ which is a function of the grammar (and not the input string), 
and where the complexity of $n \times n$ matrix multiplication is $M(n) = O(n^{\omega})$.
The parameter $d$ can be as small as $f$, meaning that we reduce
parsing complexity from $O(n^{3f})$ to $O(n^{\omega f})$, and that,
in general, the savings in the exponent is larger for more complex grammars.

LCFRS is a broad family of grammars. As such, we are able to support the findings of \citet{rajasekaran98}, who showed that tree-adjoining grammar recognition can be done in time $O(M(n^2)) = O(n^{4.76})$
(TAG can be reduced to LCFRS with $d=2$). As a result, combinatory categorial grammars, head grammars and linear indexed grammars can be recognized in time $O(M(n^2))$.
In addition, we show that inversion transduction grammars \cite[ITGs]{wu1997stochastic} can be parsed in time $O(n M(n^2)) = O(n^{5.76})$, improving the best asymptotic complexity previously known for ITGs.

\paragraph{Matrix Multiplication State of the Art} Our algorithm reduces the problem of LCFRS parsing to Boolean matrix multiplication. Let $M(n)$ be the
complexity of multiplying two such $n \times n$ matrices. These matrices can be na\"ively multiplied in $O(n^3)$ time by computing for each output cell
the dot product between the corresponding row and column in the input matrices (each such product is an $O(n)$ operation). \citet{Strassen69} discovered a
way to do the same multiplication in $O(n^{2.8704})$ time -- his algorithm is a divide and conquer algorithm that eventually uses only $7$ operations (instead of $8$) to multiply
$2 \times 2$ matrices.

With this discovery, there have been many attempts to further reduce the complexity of matrix multiplication, relying on principles similar to Strassen's method: a reduction in the number of operations it takes to multiply
sub-matrices of the original matrices to be multiplied. 
\citet{CoppersmithW87} discovered an algorithm that has the asymptotic complexity of $O(n^{2.375477})$. Others have slightly improved their algorithm, and currently there is an algorithm for matrix multiplication with $M(n) = O(n^\omega)$ 
such that $\omega = 2.3728639$ \cite{LeGall:2014}. It is known that $M(n) = \Omega(n^2 \log n)$ \cite{raz2002complexity}.

While the asymptotically best matrix multiplication algorithms 
have large constant factors lurking in the $O$-notation,
Strassen's algorithm does not, and is widely used in practice.
\citet{benedi2007fast} show speed improvement when parsing 
natural language sentences using Strassen's algorithm
as the matrix multiplication subroutine for Valiant's algorithm for CFG parsing.
This indicates that similar speed-ups may be possible in practice using our
algorithm for LCFRS parsing.

\paragraph{Main Result} Our main result is a matrix multiplication algorithm for {\bf unbalanced}, {\bf single-initial} binary LCFRS with asymptotic complexity $M(n^d) = O(n^{\omega d})$ where $d$ is the maximal number of
combination points in all grammar rules. The constant $d$ can be easily determined from the grammar at hand:
\begin{equation}
d = \max_{A \rightarrow B \, C } \max \left\{
\begin{array}{c}
\varphi(A) + \varphi(B) - \varphi(C), \\ \varphi(A) - \varphi(B) + \varphi(C), \\ -\varphi(A) + \varphi(B) + \varphi(C)
\end{array} \right\}.
\end{equation}
where $A \rightarrow B\, C$ ranges over rules in the grammar and $\varphi(A)$ is the fan-out of nonterminal $A$. 
Single-initial grammars are defined in \S\ref{section:lcfrs}, 
and include common formalisms such as tree-adjoining grammars.
Any LCFRS can be converted to single-initial form by increasing its fan-out
by at most one.
The notion of unbalanced grammars is introduced
in \S\ref{section:balanced}, and it is a condition on the set of LCFRS grammar rules that is satisfied with many practical grammars. In cases where the grammar is balanced, our algorithm
can be used as a sub-routine so that it parses the binary LCFRS in time $O(n^{\omega d + 1})$. A similar procedure was applied by \citet{nakanishi1998efficient} for multiple component context-free grammars.
See more discussion of this in \S\ref{section:mcfg}.

Our results focus on the asymptotic complexity as a function of {\em string length}. We do not give explicit grammar constants.
For other work that focuses on reducing the grammar constant in parsing, see for example \citet{eisner1999efficient,dunlop2010reducing,cohen-13a}.
For a discussion of the optimality of the grammar constants in Valiant's algorithm, see for example \citet{abboud2015if}.


\section{Background and Notation}
\label{section:lcfrs}
\label{section:notation}

This section provides background on LCFRS, and establishes notation used
in the remainder of the paper.  A reference table of notation is also 
provided in Appendix~\ref{notation}.

For an integer $n$, let $[n]$ denote the set of integers $\{ 1,\ldots, n\}$. Let $[n]_0 = [n] \cup \{ 0 \}$. For a set $X$, we denote by $X^{+}$ the set of all sequences of length 1 or more of elements from $X$.

A {\bf span} is a pair of integers denoting left and right endpoints for a substring in a larger string. The endpoints are placed in the ``spaces'' between the symbols in a string. For example, the span $(0,3)$ spans the first three
symbols in the string. For a string of length $n$, the set of potential endpoints is $[n]_0$.


We turn now to give a succinct definition for binary LCFRS. For more details about LCFRS and their relationship to other grammar formalisms, see \citet{kallmeyer-10}.
A binary LCFRS is a tuple $(\nts, \vocab, \rulessym, 
\varphi, S)$ such that:
\begin{itemize}

\item $\nts$ is the set of nonterminal symbols in the grammar.

\item $\vocab$ is the set of terminal symbols in the grammar. We assume $\nts \cap \vocab = \emptyset$.

\item $\varphi$ is a function specifying a fixed fan-out for each nonterminal ($\varphi \colon \nts \rightarrow \mathbb{N}$).


\item $\rulessym$ is a set of productions. Each production $p$ has the form $A \rightarrow g[B, C]$
where $A, B, C \in \nts$, and $g$ is a composition function 
$g : (\vocab^*)^{\varphi(B)} \times (\vocab^*)^{\varphi(C)} \rightarrow (\vocab^*)^{\varphi(A)} $,
which specifies how to assemble the $\varphi(B) + \varphi(C)$
spans of the righthand side nonterminals into the $\varphi(A)$
spans of the lefthand side nonterminal. 
We use square brackets as part of the syntax for writing productions,
and parentheses to denote the application of the function $g$.
The function $g$ must
be {\bf linear} and {\bf non-erasing}, which means
that if $g$ is applied on a pair of tuples of strings, 
then each input string appears exactly once
in the output, possibly as a {\em substring} of one of the strings in the output tuple.
Rules may also take the form $A \rightarrow g[]$,
where $g$ returns a constant tuple of one string from $\vocab$.

\item $\startsym \in \nts$ is a start symbol. Without loss of generality, we assume $\varphi(S) = 1$.

\end{itemize}
The language of an LCFRS $G = (\nts, \vocab, \rulessym, 
\varphi, S)$ is defined as follows:
 \begin{itemize}

\item We define first the set $\text{yield}(A)$ for every $A \in \nts$:
\begin{itemize}
 \item For every $A \rightarrow g[] \in \rulessym, g() \in \text{yield}(A)$.
 \item For every $A \rightarrow g[B, C] \in \rulessym$
and all tuples $\beta \in \text{yield}(B)$, $\gamma\in \text{yield}(C)$, $g(\beta,\gamma) \in \text{yield}(A)$.
 \item Nothing else is in $\text{yield}(A)$.
\end{itemize}
\item The string language of $G$ is $L(G) = \{ w \mid \langle w \rangle \in \text{yield}(S)\}$.
\end{itemize}

Intuitively, the process of generating a string from an LCFRS grammar
consists of first choosing, top-down, a production
to expand each nonterminal, and then, bottom-up, applying
the composition functions associated with each production to build the 
string.  As an example, the following context-free grammar:
\begin{align*}
S &\rightarrow A \, B \\
A &\rightarrow a \\
B &\rightarrow b
\end{align*}
corresponds to the following (binary) LCFRS:
\begin{align}
S &\rightarrow g_1[A,B] &\qquad g_1(\langle \beta_1\rangle, \langle \gamma_1 \rangle) &= \langle \beta_1\gamma_1 \rangle \label{eq:ab}\\
A &\rightarrow g_2[] &\qquad g_2() &=  \langle a \rangle \\
B &\rightarrow g_3[] &\qquad g_3() &=  \langle b \rangle 
\end{align}
The only derivation possible under this grammar
consists of the function application $g_1(g_2(), g_3()) = \langle ab \rangle$.

The following notation will be used to precisely represent the  
linear non-erasing composition functions $g$ used in a specific grammar.
For each production rule that operates on nonterminals $A$, $B$, and $C$,
we define variables from 
the set $\mathcal{S} = \{ \beta_1, \ldots, \beta_{\varphi(B)}, \gamma_1, \ldots, \gamma_{\varphi(C)} \}$.
In addition, we define variables $\alpha_{i}$ for each rule where $i \in [\varphi(A)]$, taking
values from $\mathcal{S}^+$. We write an LCFRS function as:
\[ g(\langle \beta_1, \ldots, \beta_{\varphi(B)}\rangle, 
\langle \gamma_{1}, \ldots, \gamma_{\varphi(C)} \rangle) \hfill
\hfill= \langle \alpha_{1}, \ldots, \alpha_{\varphi(A)}\rangle \]
where each $\alpha_i = \alpha_{i,1}\cdots\alpha_{i,n_i}$
specifies the parameter strings that are combined to 
form the $i$th string of the function's result tuple.
For example, for the rule in Eq.~\refeq{eq:ab}, $\alpha_{1,1} = \beta_1$
and $\alpha_{1,2} = \gamma_2$.

We adopt the following notational shorthand for LCFRS rules 
in the remainder of the paper.
We write the rule:
\begin{align*}
 &A \rightarrow g[B, C]  \\
 &g(\langle \beta_1, \ldots, \beta_{\varphi(B)}\rangle, 
\langle \gamma_{1}, \ldots, \gamma_{\varphi(C)} \rangle) \hfill
\hfill= \langle \alpha_{1}, \ldots, \alpha_{\varphi(A)}\rangle  
\end{align*}
as:
\[
 A[\alpha] \rightarrow B[\beta]\, C[\gamma]
\]
where $\alpha$ consists of a tuple of strings from
the alphabet $\{ \beta_1, \ldots, \beta_{\varphi(B)}, \gamma_1, \ldots, \gamma_{\varphi(C)} \}$.  In this notation, $\beta$ is always the tuple
$\langle \beta_1, \ldots, \beta_{\varphi(B)}\rangle$,
and $\gamma$ is always $\langle \gamma_1, \ldots, \gamma_{\varphi(C)}\rangle$.
We include $\beta$ and $\gamma$ in the rule notation merely
to remind the reader of the meaning of the symbols in $\alpha$.

For example, with context-free grammars, rules have the 
form:
\[
 A[\langle\beta_1\gamma_1\rangle] \rightarrow B[\langle\beta_1\rangle]\, C[\langle\gamma_1\rangle]
\]
indicating that $B$ and $C$ each have one span, and are concatenated in order
to form $A$. 

\begin{figure}
\begin{center}
\begin{tikzpicture}
\draw(-1.5, 1.25)node{$B$};
\draw(-1.5, 2.25)node{$C$};
\draw(-1.5, 0.25)node{$A$};

\draw[fill=azure](0, 1)rectangle(1, 1.5); \draw(0.5, 1.25)node{$\beta_1$};
\draw[fill=azure](4, 1)rectangle(5, 1.5); \draw(4.5, 1.25)node{$\beta_2$};
\draw[fill=azure](1, 2)rectangle(2, 2.5); \draw(1.5, 2.25)node{$\gamma_1$};
\draw[fill=azure](3, 2)rectangle(4, 2.5); \draw(3.5, 2.25)node{$\gamma_2$};
\draw[fill=azure](-.2, 0.75)rectangle(5.2, 0.75); 
\draw[fill=azure](0, 0)rectangle(2, 0.5); \draw(1, 0.25)node{$\beta_1\gamma_1$};
\draw[fill=azure](3, 0)rectangle(5, 0.5); \draw(4, 0.25)node{$\gamma_2\beta_2$};
\end{tikzpicture} 
\end{center}
\caption{An example of a combination of spans for Tree Adjoining Grammars (TAG) for the adjunction operation in terms of binary LCFRS.
The rule in Eq.~\protect\refeq{eq:tag} specifies how two nonterminals $B$ and $C$ are combined together into a nonterminal $A$.\label{fig:tag}}
\end{figure}

A binary tree-adjoining grammar can also be represented as a binary LCFRS \cite{vw94}.
Figure~\ref{fig:tag} demonstrates how the adjunction operation is done with binary LCFRS. Each gray block denotes a span, and the adjunction operator takes
the first span of nonterminal $B$ and concatenates it to the first span of nonterminal $C$ (to get the first span of $A$), and then takes the second span
of $C$ and concatenates it with the second span of $B$ (to get the second span of $A$).
For tree-adjoining grammars, rules have the form:
\begin{equation}
A[\langle\beta_1\gamma_1, \gamma_2\beta_2\rangle] \rightarrow B[\langle\beta_1, \beta_2\rangle]\, C[\langle\gamma_1, \gamma_2\rangle] \label{eq:tag}
\end{equation}

The {\bf fan-out} of a nonterminal is the number of spans in the input
sentence that it covers.  The fan-out of CFG rules is one, and the fan-out 
of TAG rules is two.
The fan-out of the grammar, $f$, is the maximum 
fan-out of its nonterminals:
\begin{equation}
f = \max_{A \in \nts} \varphi(A).\label{eq:f}
\end{equation}

We sometimes refer to the {\bf skeleton} of a grammar rule $A[\alpha] \rightarrow B[\beta] \, C[\gamma]$, which is just the context-free rule $A \rightarrow B\, C$, omitting the variables.
In that context, a logical statement such as $A \rightarrow B\, C \in \rulessym$ is true if there is any rule $A[\alpha] \rightarrow B[\beta] \, C[\gamma] \in \rulessym$ with some $\alpha,\beta$, and $\gamma$.

For our parsing algorithm, we assume that the grammar is in a
normal form such that the variables $\beta_1, \ldots, \beta_{\varphi(B)}$
appear in order in $\alpha$, that is, that the spans of $B$ are
not re-ordered by the rule, and similarly we assume
that $\gamma_1, \ldots, \gamma_{\varphi(C)}$ appear in order.  If
this is not the case in some rule, the grammar can be transformed
by introducing a new nonterminal for each permutation of a
nonterminal that can be produced by the grammar.
We further
assume that  $\alpha_{1,1} = \beta_1$, that
is, that the first span of $A$ begins with material produced by
$B$ rather than by $C$.  If this not the case for some rule, $B$
and $C$ can be exchanged to satisfy this condition.

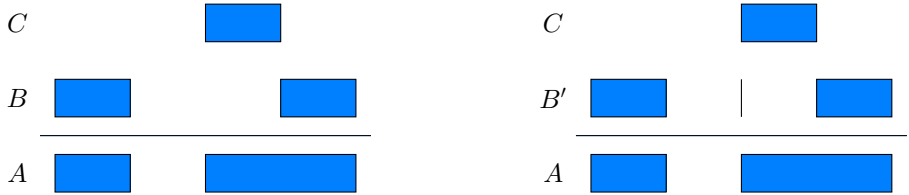
\begin{figure}
\begin{tabular}{lll}
\begin{tikzpicture}
\draw(-0.5, 1.25)node{$B$};
\draw(-0.5, 2.25)node{$C$};
\draw(-0.5, 0.25)node{$A$};

\draw[fill=azure](0, 1)rectangle(1, 1.5);
\draw[fill=azure](3, 1)rectangle(4, 1.5);
\draw[fill=azure](2, 2)rectangle(3, 2.5);
\draw[fill=azure](-.2, 0.75)rectangle(4.2, 0.75); 
\draw[fill=azure](0, 0)rectangle(1, 0.5);
\draw[fill=azure](2, 0)rectangle(4, 0.5);
\end{tikzpicture} 
&
\hspace{0.5in}
&
\begin{tikzpicture}
\draw(-0.5, 1.25)node{$B'$};
\draw(-0.5, 2.25)node{$C$};
\draw(-0.5, 0.25)node{$A$};

\draw[fill=azure](0, 1)rectangle(1, 1.5);
\draw[fill=azure](2, 1)rectangle(2, 1.5); 
\draw[fill=azure](3, 1)rectangle(4, 1.5);
\draw[fill=azure](2, 2)rectangle(3, 2.5);
\draw[fill=azure](-.2, 0.75)rectangle(4.2, 0.75); 
\draw[fill=azure](0, 0)rectangle(1, 0.5);
\draw[fill=azure](2, 0)rectangle(4, 0.5);
\end{tikzpicture} 
\end{tabular}
\caption{Conversion of a dual-initial rule to a single-initial rule.}\label{fig:single-initial}
\end{figure}

We refer to an LCFRS rule $A \rightarrow B \, C$
as {\bf single-initial} if the leftmost endpoint of $C$ is internal
to a span of $A$, and {\bf dual-initial} if the leftmost endpoint
of $C$ is the beginning of a span of $A$.  Our algorithm will require 
the input LCFRS to be in single-initial form, meaning
that all rules are single-initial.  We note that 
grammars for common formalisms including TAG and synchronous
context-free grammar (SCFG) are in this form.
If a grammar is not in single-initial form, dual-initial rules
can converted to single-initial form by adding a empty span to $B$
which combines with the first spans of $C$ immediately to its left,
as shown in Figure~\ref{fig:single-initial}.
Specifically, for each dual-initial
rule $A \rightarrow B \, C$, if the first 
span of $C$ appears between spans $i$ and $i+1$ of $B$, 
create a new nonterminal $B'$ with $\varphi(B') = \varphi(B) +1$, and
add a rule $B'\rightarrow B$, where $B'$ produces $B$ along with
a span of length zero between spans $i$ and $i+1$ of $B$.
We then replace the rule $A \rightarrow B \, C$
with $A \rightarrow B' \, C$, where the new span of $B'$
combines with $C$ immediately to the left of $C$'s first span.
Because the new nonterminal $B'$ has fan-out one greater
than $B$, this grammar transformation can increase a grammar's
fan-out by at most one.




By limiting ourselves to binary LCFRS grammars, we do not necessarily restrict the power of our results. Any LCFRS with arbitrary rank (i.e.
with an arbitrary number of nonterminals in the right-hand side) can be converted to a binary LCFRS (with potentially a larger fan-out). See discussion
in \S\ref{section:binarization-strategy}.

\begin{example}
Consider the phenomenon of cross-serial dependencies that exists in certain languages.
It has been used in the past \cite{Shieber85} to argue that Swiss-German is not
context-free. One can show that there is a homomorphism between Swiss-German and the alphabet $\{ a,b,c,d\}$ such that the image
of the homomorphism intersected with the regular language $a^{\ast}b^{\ast}c^{\ast}d^{\ast}$ gives the language $L = \{ a^m b^n c^m d^n \mid m,n \ge 1 \}$.
Since $L$ is not context-free, this implies that Swiss-German is not context-free, because context-free languages
are closed under intersection with regular languages.

Tree-adjoining grammars, on the other hand, are mildly context-sensitive formalisms that can handle such cross-serial dependencies
in languages (where the $a$s are aligned with $c$s and the $b$s are aligned with the $d$s). For example, a tree-adjoining grammar for generating $L$ would
include the following initial and auxiliary trees (nodes marked by $\ast$ are nodes where adjunction is not allowed):

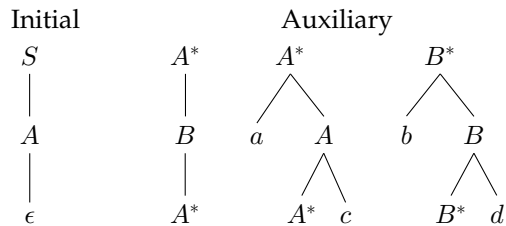
\begin{center}
\begin{tabular}{lllll}
\multicolumn{1}{c}{Initial} & $\,\,\,\,$ & \multicolumn{3}{c}{Auxiliary} \\
\Tree[.$\startsym$ [.$A$ [.$\epsilon$ ] ] ] &  & \Tree[.$A^{\ast}$ [.$B$ [.$A^\ast$ ] ] ] & \Tree[.$A^{\ast}$ [.$a$ ] [.$A$ [.$A^\ast$ ] [.$c$ ] ] ] & \Tree[.$B^{\ast}$ [.$b$ ] [.$B$ [.$B^\ast$ ] [.$d$ ] ] ]
\end{tabular}
\end{center}

This TAG corresponds to the following LCFRS:
\begin{align}
\startsym &\rightarrow g_1[A] &\qquad g_1(\langle \beta_1,\beta_2\rangle) &= \langle \beta_1\beta_2 \rangle \\
A &\rightarrow g_4[B] &\qquad g_4(\langle \beta_1,\beta_2\rangle) &= \langle \beta_1,\, \beta_2 \rangle \\
A &\rightarrow g_2[A] &\qquad g_2(\langle \beta_1,\beta_2\rangle) &= \langle a\beta_1,\, c\beta_2 \rangle \\
B &\rightarrow g_5[B] &\qquad g_5(\langle \beta_1,\beta_2\rangle) &= \langle b\beta_1,\, d\beta_2 \rangle \\
A &\rightarrow g_3[] &\qquad g_3() &=  \langle \varepsilon,\, \varepsilon \rangle \\
B &\rightarrow g_6[] &\qquad g_6() &= \langle \varepsilon,\, \varepsilon \rangle
\end{align}
Here we have one unary LCFRS rule for the initial tree, 
one unary rule for each adjunction tree, and one null-ary rule for each nonterminal 
producing a tuple of empty strings in order to represent TAG tree nodes
at which no adjunction occurs. 
The LCFRS given above does not satisfy our normal form
requiring each rule to have either two nonterminals on the righthand side
with no terminals in the composition function, or zero nonterminals 
with a composition function returning fixed strings of terminals.
However, it can be converted to such a form through a process
analogous to converting a CFG to Chomsky Normal Form.
For adjunction trees, the two strings returned by the composition function
correspond the the material to the left and right of the foot node.
The composition function merges terminals at the leaves of the adjunction tree
with material produced by internal nodes of the tree at which adjunction 
may occur.

In general, binary LCFRS are more expressive than TAGs
because they can have nonterminals with fan-out greater than two, 
and because they can interleave the arguments of the composition function
in any order.

\end{example}

\section{A Sketch of the Algorithm}
\label{section:sketch}

Our algorithm for LCFRS string recognition is inspired by the algorithm of \citet{valiant75}.
It introduces a few important novelties that make it possible to use matrix multiplication for the goal of LCFRS recognition.

The algorithm relies on the observation that it is possible to construct a matrix $T$ with a specific non-associative multiplication and addition operator
such that multiplying $T$ by itself $k$ times on the left or on the right yields $k$-step derivations for a given string. The row and column indices of the matrix together assemble
a set of spans in the string (the fan-out of the grammar determines the number of spans). 
Each cell in the matrix keeps track of the nonterminals
that can dominate these spans. Therefore, computing the transitive closure of this matrix yields in each matrix cell the set of nonterminals that can dominate the assembled indices' spans
for the specific string at hand.

There are several key differences between Valiant's algorithm and our algorithm. Valiant's algorithm has a rather simple matrix indexing scheme for the matrix: the rows correspond
to the left endpoints of a span and the columns correspond to its right endpoints. Our matrix indexing scheme can mix both left endpoints and right endpoints at either the
rows or the columns. This is necessary because with LCFRS, spans for the right-hand side of an LCFRS rule can combine in various ways into a new set of spans for the left-hand side.

\begin{figure}
\begin{center}
\begin{tabular}{p{3.0in}p{1.0in}}
{
\begin{tikzpicture}[baseline]
\draw(-1.5, 1.25)node{$B$};
\draw(-1.5, 2.25)node{$C$};
\draw(-1.5, 0.25)node{$A$};
\draw[fill=azure](0, 1)rectangle(1, 1.5); \draw(0.5, 1.25)node{$\beta_1$};
\draw[fill=azure](4, 1)rectangle(5, 1.5); \draw(4.5, 1.25)node{$\beta_2$};
\draw[fill=azure](1, 2)rectangle(2, 2.5); \draw(1.5, 2.25)node{$\gamma_1$};
\draw[fill=azure](3, 2)rectangle(4, 2.5); \draw(3.5, 2.25)node{$\gamma_2$};
\draw[fill=azure](-.2, 0.75)rectangle(5.2, 0.75); 
\draw[fill=azure](0, 0)rectangle(2, 0.5); \draw(1, 0.25)node{$\beta_1\gamma_1$};
\draw[fill=azure](3, 0)rectangle(5, 0.5); \draw(4, 0.25)node{$\gamma_2\beta_2$};
\draw[dotted](0,2.5)--(0,-0.25); \draw(0,-.5)node{$i$};
\draw[dotted](1,2.5)--(1,-0.25); \draw(1,-.5)node{$j$};
\draw[dotted](2,2.5)--(2,-0.25); \draw(2,-.5)node{$k$};
\draw[dotted](3,2.5)--(3,-0.25); \draw(3,-.5)node{$\ell$};
\draw[dotted](4,2.5)--(4,-0.25); \draw(4,-.5)node{$m$};
\draw[dotted](5,2.5)--(5,-0.25); \draw(5,-.5)node{$n$};
\draw(0,-.85)node{$1$};
\draw(1,-.85)node{$2$};
\draw(2,-.85)node{$4$};
\draw(3,-.85)node{$5$};
\draw(4,-.85)node{$7$};
\draw(5,-.85)node{$8$};
\end{tikzpicture} 
}
&
\begin{tabular}{l}
$ [ C, j, k, \ell, m ]$\\
$ [ B, i, j, m, n ]$\\
\hline
$ [ A, i, k, \ell, n ]$\\
\end{tabular}
\end{tabular}
\end{center}
\caption{A demonstration of a parsing step for the combination of spans in Figure~\ref{fig:tag}.
During parsing, the endpoints of each span are instantiated 
with indices into the string.  The variables for these
indices shown on the left correspond to the
logical induction rule on the right.
The specific choice of indices shown at the bottom
is used in our matrix multiplication example in \S\ref{section:sketch}.}\label{fig:tag-indices}
\end{figure}
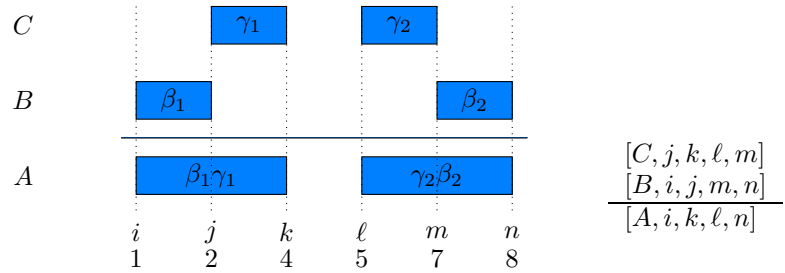

In addition, our indexing scheme is ``over-complete.'' This means that different cells in the matrix $T$ (or its matrix powers) are equivalent and should consist of the
same nonterminals. The reason we need such an over-complete scheme is again because of the possible ways spans of a right-hand side can combine in an LCFRS. To address this over-completeness,
we introduce into the multiplication operator a ``copy operation'' that copies nonterminals between cells in order to maintain the same set of nonterminals in equivalent cells.

To give a preliminary example, consider the tree-adjoining grammar
rule shown in Figure~\ref{fig:tag}.
We consider an application of the rule with the endpoints of 
each span instantiated as shown in Figure~\ref{fig:tag-indices}.
With our algorithm, this operation will translate into the following sequence of matrix transformations.
We will start with the following matrices, $T_1$ and $T_2$:

$\,$

\begin{tabular}{ll}
\begin{minipage}{0.5\textwidth}
$
\bordermatrix{T_1 & &  &  (2,7) & & & \cr
                 &   &  &  &  &   &  &  \cr
                &   &  &  &    &   & &  \cr
                (1,8) &  &   &  \{ \ldots, B, \ldots \} &  &  &   &  \cr
                 &  &   &  &  &    &   &   \cr
                 &   & &  &  &  &    &   \cr
                &   &  &  & &  &   &  \cr
                &  &  & & & &    &  \cr
                 &   &  &  &  &  &  & }
$
\end{minipage}

&

\begin{minipage}{0.5\textwidth}
$
\bordermatrix{T_2 &  &  &   & (4,5) & & \cr
                 &   &  &  &  &   &  &  \cr
                &   &  &  &    &   & &  \cr
                 &  &   &   &  &  &   &  \cr
                (2,7) &  &   &  & \{ \ldots, C, \ldots \} &    &   &   \cr
                 &   & &  &  &  &    &   \cr
                &   &  &  & &  &   &  \cr
                &  &  & & & &    &  \cr
                 &   &  &  &  &  &  & }.
$
\end{minipage}

\end{tabular}

$\,$

For $T_1$, for example, the fact that $B$ appears for the pair of addresses $(1,8)$ (for row) and $(2,7)$ for column denotes
that $B$ spans the constituents $(1,2)$ and $(7,8)$ in the string (this is assumed to be true -- in practice, it is the result
of a previous step of matrix multiplication).  Similarly, with $T_2$, $C$ spans the constituents $(2,4)$ and $(5,7)$.

Note that $(2,7)$ are the two positions in the string 
where $B$ and $C$ meet, and that because $B$ and $C$ 
share these two endpoints, they can combine to form $A$.   
In the matrix representation,  
$(2,7)$ appears as the column address of $B$ and as 
the row address of $C$, meaning that $B$ and $C$ appear in
cells that are combined during matrix multiplication.
The result of multiplying $T_1$ by $T_2$ is the following:

$$
\bordermatrix{T_1 T_2 &  &  &   & (4,5) & & \cr
                 &   &  &  &  &   &  &  \cr
                &   &  &  &    &   & &  \cr
                (1,8) &  &   &  & \{ \ldots, A, \ldots \} &    &   &   \cr
                 &  &   &   &  &  &   &  \cr
                 &   & &  &  &  &    &   \cr
                &   &  &  & &  &   &  \cr
                &  &  & & & &    &  \cr
                 &   &  &  &  &  &  & }.
$$

Now $A$ appears in the cell that corresponds to the spans $(1,4)$ and $(5,8)$. This is the result of merging
the spans $(1,2)$ with $(2,4)$  (left span of $B$ and left span of $C$) into $(1,4)$ and the merging
of the spans $(5,7)$ and $(7,8)$ (right span of $C$ and right span of $B$) into $(5,8)$. Finally, an additional
copying operation will lead to the following matrix:

$$
\bordermatrix{T_3 &  &  &  & & (5,8)  & \cr
                 &   &  &  &  &   &  &  \cr
                (1,4) &   &  &  &    & \{ \ldots, A, \ldots \}  & &  \cr
                 &  &   &  &  &    &   &   \cr
                 &  &   &   &  &  &   &  \cr
                 &   & &  &  &  &    &   \cr
                &   &  &  & &  &   &  \cr
                &  &  & & & &    &  \cr
                 &   &  &  &  &  &  & }.
$$

Here, we copy the nonterminal $A$ from the address with the row $(1,8)$ and column $(4,5)$ into
the address with the row $(1,4)$ and column $(5,8)$. Both of these addresses correspond to
the same spans $(1,4)$ and $(5,8)$. Note that matrix row and column
addresses can mix both starting points
of spans and ending points of spans.

\section{A Matrix Multiplication Algorithm for LCFRS}

We turn next to give a description of the algorithm. Our description is constructed as follows:

\begin{itemize}

\item In \S\ref{section:matrix-structure} we describe the basic matrix structure which is used for LCFRS recognition. This construction depends on a parameter $d$, the contact rank, which is a function
of the underlying LCFRS grammar we parse with. We also describe how to create a seed matrix, for which we need to compute the transitive closure.

\item In \S\ref{section:mul} we define the multiplication operator between cells of the matrices we use. This multiplication operator is distributive, but not associative, and as such,
we use Valiant's specialized transitive closure algorithm to compute transitive closure of the seed matrix given a string.

\item In \S\ref{sec:d} we define the contact rank parameter $d$. The smaller $d$ is, the more efficient it is to parse with the specific grammar.

\item In \S\ref{section:balanced} we define when a binary LCFRS is ``balanced.'' This is an end case that increases the final complexity of our algorithm by a factor of $O(n)$.
Nevertheless, it is an important end case that
appears in applications, such as inversion transduction grammars.

\item In \S\ref{section:transitive-closure} we tie things up, and show that computing the transitive closure of the seed matrix we define in \S\ref{section:matrix-structure} yields
a recognition algorithm for LCFRS.

\end{itemize}

\subsection{Matrix Structure}
\label{section:matrix-structure}

The algorithm will seek to compute the transitive closure of a seed matrix $T(d)$, where $d$ is a constant determined by the grammar (see \S\ref{sec:d}).
The matrix rows and columns are indexed by the set $N(d)$ defined as:
\begin{equation}
N(d) =  \bigcup_{d'=1}^d ( [n]_0 \times \{0,1\} )^{d'},\label{eq:nd}
\end{equation}
where $n$ denotes the length of the sentence,
and the exponent $d'$ denotes a repeated Cartesian product.
Thus each element of $N(d)$ is a sequence of indices
into the string, where each index is annotated with a bit 
(an element of the set $\{0,1\}$) indicating whether it 
is {\bf marked} or unmarked.  Marked indices will
be used in the copy operator defined later.  Indices are
unmarked unless specified as marked: we use 
$\hat x$ to denote a marked index $(x,1)$ with $x\in[n]_0$.
 
In the following, it will be safe to assume sequences from $N(d)$
are monotonically increasing in their indices. For an $i \in N(d)$, we overload notation, and often refer
to the set of all elements in the first coordinate of each element in the sequence (ignoring the additional bits). As such,

\begin{itemize}

\item The set $i \cup j$ is defined for $j \in N(d)$. 

\item If we state that $i$ is in $N(d)$ and includes a {\em set} of endpoints, it means that $i$ is a sequence of these integers (ordered lexicographically) with the bit
part determined as explained in the context (for example, all unmarked).

\item The quantity $|i|$ denotes the length of the sequence.

\item The quantity $\min i$ denotes the smallest index among the first coordinates of all elements in the sequence $i$ (ignoring the additional bits).

\end{itemize}

We emphasize that the variables $i$, $j$, and $k$ are mostly elements in $N(d)$ as overloaded above,
not integers, throughout this paper; we choose the symbols $i$, $j$, and $k$
by analogy to the variables in the CKY parsing algorithm, and also
because we use the sequences as addresses for matrix rows and columns.
For $i,j \in N(d)$, we define $m(i,j)$ to be the set of $f' = \displaystyle\frac{1}{2} | i \cup j |$ pairs $\{ (\ell_1,\ell_2), (\ell_3,\ell_4), \ldots, (\ell_{2f'-1},\ell_{2f'}) \}$
such that $\ell_{k} < \ell_{k+1}$ for $k \in [2f'-1]$ and $(\ell_k, 0) \in i \cup j$ for $k \in [2f']$. 
This means that $m(i,j)$ takes as input the two sequences in matrix indices, merges them, sorts them, then divides this sorted list into a set of $f'$ consecutive pairs.
Whenever $\min j \le \min i$,  $m(i,j)$ is undefined. The interpretation
of this is that $\ell_1$ should always belong to $i$ and not $j$. See more details in \S\ref{section:upper-tri}. 
In addition, if any element of $i$ or $j$ is marked, $m(i,j)$ is undefined.

We define an order $\ijleq$  on elements $i$ and $j$ of $N(d)$ by
first sorting the sequences $i$ and $j$ and then comparing $i$ and $j$
lexicographically (ignoring the bits). 
This ensures that $i \ijleq j$ if $\min i < \min j$.
We assume that the rows and columns of our matrices are arranged in this order.
For the rest of the discussion, we assume that $d$ is a constant, and refer to $T(d)$ as $T$ and $N(d)$ as $N$.

We also define the set of triples $M$ as the following Cartesian product:
\begin{equation}
M = (\nts \cup \{ \frow, \fcol, \trow, \tcol, \urow, \ucol \}) \times N \times N,
\end{equation}
where $\frow$, $\tcol$, $\fcol$, $\trow$, $\urow$, and $\ucol$ are six
special pre-defined symbols.\footnote{The symbols will be used for ``copying commands:'' (1) ''from row'' ($\frow$); (2) ``from column'' ($\fcol$); (3) ``to row'' ($\trow$);
(4) ``to column'' ($\tcol$); (5) ``unmark row'' ($\urow$); (6) ``unmark column'' ($\ucol$).} Each cell $T_{ij}$ in $T$ is a set
such that $T_{ij} \subset M$.

The intuition behind matrices of the type of $T$ (meaning $T$ and, as we see later, products of $T$ with itself, or its transitive closure) is that each cell indexed by $(i,j$) in such
a matrix consists of all nonterminals that can be generated by the grammar when parsing a sentence such that these nonterminals span the constituents $m(i,j)$ (whenever $m(i,j)$ is defined).
Our normal form for LCFRS ensures that spans
of a nonterminal are never re-ordered, meaning that it is not necessary to 
retain information about which indices demarcate which components of 
the nonterminal, because one can sort the indices and take
the first two indices as delimiting the first span, the
second two indices as delimiting the second span, and so on.
The two additional $N$ elements in each triplet in a cell are actually just copies of the row and column indices of that cell. As such, they are identical for all triplets in that cell.
The additional $\frow$, $\tcol$, $\fcol$, $\trow$, $\urow$, $\ucol$ symbols 
are symbols that indicate to the matrix multiplication operator that a ``copying operation'' should happen between equivalent cells (\S\ref{section:mul}).

Figure~\ref{fig:seed-matrix} gives an algorithm to seed the
initial matrix $T$.  Entries added in step 2 of the algorithm
correspond to entries in the LCFRS parsing chart that can be
derived immediately from terminals in the string.
Entries added in step 3 of the algorithm do not depend 
on the input string or input grammar, but rather initialize elements
used in the copy operation described in detail in \S\ref{section:mul}.
Because the algorithm only initializes entries with $i<j$,
the matrix $T$ is guaranteed to be upper triangular, a fact
which we will take advantage of in \S\ref{section:mul}.

\begin{figure}

\framebox{\parbox{\columnwidth}{

{\bf Inputs:} An LCFRS grammar as defined in \S\ref{section:lcfrs} and a sentence $w_1 \cdots w_n$. \\

{\bf Outputs:} A seed matrix $T$ with rows and columns indexed by $N$, such that each cell in $T$ is a subset of $M$ \\

{\bf Algorithm:}

\begin{enumerate}

\item Set $T_{ij} = \emptyset$ for all $i,j \in N$.

\item For each $i,j \in N$, for each nonterminal $A \in \nts$, set $T_{ij} \leftarrow T_{ij} \cup \{ (A, i, j ) \}$ if $m(i,j) = \{ (\ell_1,\ell_2), (\ell_3,\ell_4), \ldots, (\ell_{2f-1},\ell_{2f}) \}$
and 
%
%
%
there is a rule in the grammar $A \rightarrow g(), g() = \langle \alpha_1 , \ldots, \alpha_{\varphi(A)}\rangle$, where  $\alpha_i = \alpha_{i,1}\cdots\alpha_{i,n_i}$,
and for each $i$ and $j$,
$\alpha_{i,j} = w_{\ell_{2i-1}+j}$ (i.e. $\alpha_{i,j}$ is the
$(\ell_{2i-1}+j)$th word in the sentence).


\item For each $i,j \in N$ such that $i \ijleq j$
\begin{enumerate}[label=\alph*.]
\item  $T_{ij} \leftarrow T_{ij} \cup \{ ( \tcol, i, j ) \}$ if all indices in $i$ are unmarked and $j = \insertt(i,\hat x)$ for some $x$
\item  $T_{ij} \leftarrow T_{ij} \cup \{ ( \frow, i, j ) \}$ if all indices in $j$ are unmarked and $i = \remove(j, x)$ for some $x$
\item  $T_{ij} \leftarrow T_{ij} \cup \{ ( \urow, i, j ) \}$ if all indices in $i$ are unmarked and $i = \insertt(\remove(j,\hat x),  x)$ for some $x$
\item  $T_{ij} \leftarrow T_{ij} \cup \{ ( \trow, i, j ) \}$ if all indices in $j$ are unmarked and $i = \insertt(j, \hat x)$ for some $x$
\item  $T_{ij} \leftarrow T_{ij} \cup \{ ( \fcol, i, j ) \}$ if all indices in $i$ are unmarked and $j = \remove(i, x)$ for some $x$
\item  $T_{ij} \leftarrow T_{ij} \cup \{ ( \ucol, i, j ) \}$ if all indices in $j$ are unmarked and $j = \insertt(\remove(i, \hat x), x)$ for some $x$
\end{enumerate}

\end{enumerate}

}}

\caption{\label{fig:seed-matrix} An algorithm for computing the seed matrix $T$.  The function $\remove(v,x)$ takes a sequence of integers $v$ and removes $x$ from it, if it is in there. The function
$\insertt(v,x)$ takes a sequence of integers and adds $x$ to it.}
\end{figure}

\subsubsection{Configurations}
\label{sec:config}
Our matrix representation requires that a nonterminal appears in more
than one equivalent cell in the matrix, and the specific set of cells
required depends on the specific patterns in which spans are combined
in the LCFRS grammar.  We now present a precise description of these
cells by defining the {\bf configuration} of a nonterminal in a rule.
The concept of a configuration is designed to represent which endpoints of 
spans of the rule's righthand side (r.h.s.)\ nonterminals $B$ and $C$ meet one
another to produce larger spans, and which endpoints, on the other hand,
become endpoints of spans of the lefthand side (l.h.s.)\ nonterminal $A$.

For each of the three nonterminals involved in a rule, 
the configuration is the set of endpoints in the row address 
of the nonterminal's matrix cell.  To make this precise,
for a nonterminal $B$ with fan-out $\varphi(B)$, we number the endpoints
of spans with integers in the range $1$ to $2\varphi(B)$.
In a rule $A[\alpha] \rightarrow B[\beta] \, C[\gamma]$,
the configuration of $B$ is the subset of $[2\varphi(B)]$
of endpoints of $B$ that do not combine with endpoints of $C$ in order
to form a single span of $A$.  The endpoints will form the row address 
for $B$.
Formally, let $\beta = \langle \beta_1, \ldots,\beta_{\varphi(B)}\rangle$,
and let $\alpha = \langle \alpha_{1,1}\cdots\alpha_{1,n_1},\ldots,\alpha_{\varphi(A),1}\cdots\alpha_{\varphi(A),n_{\varphi(A)}}\rangle$.
Then the set of non-combination endpoints of $B$ is defined as:
\[
\configtwo(r) = \{ 2i \mid \beta_i = \alpha_{j,n_j} \text{ for some } j \} \cup
    \{ 2i-1 \mid \beta_{i} = \alpha_{j,1} \text{ for some } j \} 
\]
where the first set defines right ends of spans of $B$ that are right ends of some span of $A$,
and the second set defines left ends of spans of $B$ that are left ends of some span of $A$.
For example, given that CFG rules have the from 
\[
r = A[\langle\beta_1\gamma_1\rangle] \rightarrow B[\langle\beta_1\rangle]\, C[\langle\gamma_1\rangle]
\]
the configuration $\configtwo(r)$ is $\{ 1 \}$ because, of $B$'s two endpoints,
only the first is also an endpoint of $A$.
For the TAG rule $t$ shown in Figure~\ref{fig:tag},
 $\configtwo(t) = \{ 1, 4 \}$ because, of $B$'s four
endpoints, the first and fourth are also endpoints of $A$.

For the second r.h.s.\ nonterminal of a rule $r$, 
the configuration consists of the set of endpoints in the 
row address for $C$, which are the endpoints that {\em do} combine with $B$:
\[
\configthree(r) = \{ 2i \mid \gamma_{i} = \alpha_{j,k} \text{ for some } 1 \le k < n_j \} \cup
    \{ 2i-1 \mid \gamma_{i} = \alpha_{j,k} \text{ for some } 1 < k \le n_j \} 
\]
where the first set defines right ends of spans of $C$ that are internal to some span of $A$,
and the second set defines lefts ends of spans of $C$ that are internal to some span of $A$.
For example, any CFG rule $r$ has configuration, $\configthree(r) = \{ 1 \}$,
because the first endpoint of $C$ is internal to $A$. 
For the TAG rule $t$ shown in Figure~\ref{fig:tag},
 $\configthree(t) = \{ 1, 4 \}$ because, of $C$'s four
endpoints, the first and fourth are internal $A$.

For the l.h.s.\ nonterminal $A$ of the rule, matrix multiplication 
will produce an entry in the matrix cell where the row address corresponds
to the endpoints from $B$, and the column address corresponds 
to the endpoints from $C$.  To capture this
partition of the endpoints of $A$, we define
\[
\configone(r) = \{ 2i \mid \alpha_{i,n_i} = \beta_j \text{ for some } j \} \cup
\{ 2i-1 \mid \alpha_{i,1} = \beta_j \text{ for some } j \},
\]
where the first set defines right ends of spans of $A$ that are formed from $B$,
and the second set defines left ends of spans of $A$ that are formed from $B$.
For example, any CFG rule $r$ has configuration, $\configone(r) = \{ 1 \}$,
because only the first endpoint of $A$ is derived from $B$. 
For the TAG rule $t$ shown in Figure~\ref{fig:tag},
 $\configone(t) = \{ 1, 4 \}$ because, of $A$'s four
endpoints, the first and fourth are derived from $B$.

\subsection{Definition of Multiplication Operator}
\label{section:mul}
\label{section:upper-tri}

We need to define a multiplication operator $\otimes$ between a pair of elements $R,S \subset M$. Such a multiplication operator induces multiplication between matrices of the type of $T$, just by
defining for two such matrices, $T_1$ and $T_2$, a new matrix of the same size $T_1 \otimes T_2$ such that:
\begin{equation}
[T_1 \otimes T_2]_{ij} = \bigcup_{k \in N} \left( [T_1]_{ik} \otimes [T_2]_{kj} \right), \label{eq:a}
\end{equation}
We also use the $\cup$ symbol to denote coordinate-wise union of cells in the matrices it operates on.

\begin{figure}

\framebox{\parbox{\columnwidth}{
\raggedright
{\bf Inputs:} A pair of elements $R,S \subset M$. \\

{\bf Outputs:} A new subset of $M$, denoted by $(R \otimes S)$. \\

{\bf Algorithm:}

\begin{enumerate}[label=\arabic*.,ref=\arabic*]

\item $(R \otimes S) = \emptyset$.

\item \label{step:ab} For each pair of elements $r = (B, i, k)$ and $s = (C, k, j)$ where $r \in R$ and $s \in S$ add
an element $(A, i, j)$ to $(R \otimes S)$ if:

\begin{enumerate}[label=\alph*.,ref=\alph*]

\item There is a binary rule in the LCFRS grammar $r = A[\alpha] \rightarrow B[\beta] \, C[\gamma]$.

\item All indices in $i$, $j$, and $k$ are unmarked.

\item \label{step:bconfig} Configuration of $B$:
assume $m(i,k) = \{ (\ell_1,\ell_2), \ldots, (\ell_{2\varphi(B)-1},\ell_{2\varphi(B)}) \}$.\\
We require that $i = \{ \ell_q \mid q \in \configtwo(r) \}$

\item \label{step:cconfig} Configuration of $C$:
assume $m(k,j) = \{ (\ell'_1,\ell'_2), \ldots, (\ell'_{2\varphi(C)-1},\ell'_{2\varphi(C)}) \}$.\\
We require that $k = \{ \ell'_q \mid q \in \configthree(r) \}$

\item \label{step:aconfig} Configuration of $A$:
assume $m(i,j) = \{ (\ell''_1,\ell''_2), \ldots, (\ell''_{2\varphi(A)-1},\ell''_{2\varphi(A)}) \}$.\\
We require that $i = \{ \ell''_q \mid q \in \configone(r) \}$

\end{enumerate}

\item For each pair of elements $r = (A,i,k)$ and $s = (\tcol, k, j)$, add the element $(A,i,j)$ to $(R \otimes S)$ if:
$i$ contains only unmarked indices, and
$\hat x \in j$ and $x \in i$ for some $x$.

\item For each pair of elements $r = (\frow, i, k)$ and $s = (A, k, j)$, add the element $(A,i,j)$ to $(R \otimes S)$ if:
$\hat x \in j$ and $x \not\in i$ and for some $x$.

\item For each pair of elements $r = (A,i,k)$ and $s = (\ucol, k, j)$, add the element $(A,i,j)$ to $(R \otimes S)$ if:
$|i \cup j| = 2\varphi(A)$.

\item For each pair of elements $r = (\trow,i,k)$ and $s = (A, k, j)$, add the element $(A,i,j)$ to $(R \otimes S)$ if:
$j$ contains only unmarked indices, and
$\hat x \in i$ and $x \in j$ for some $x$.

\item For each pair of elements $r = (A, i, k)$ and $s = (\fcol, k, j)$, add the element $(A,i,j)$ to $(R \otimes S)$ if:
$\hat x \in i$ and $x \not\in j$ for some $x$.

\item For each pair of elements $r = (\urow,i,k)$ and $s = (A, k, j)$, add the element $(A,i,j)$ to $(R \otimes S)$ if:
$|i \cup j| = 2\varphi(A)$.

\end{enumerate}

}}

\caption{\label{fig:otimes} An algorithm for the product of two matrix elements.}
\end{figure}

The operator $\otimes$ we define is not associative, but it is distributive over $\cup$. This means that for $R,S_1,S_2 \subset M$ it holds that:
\begin{equation}
R \otimes (S_1 \cup S_2) = (R \otimes S_1) \cup (R \otimes S_2).
\end{equation}

In addition, whenever $R = \emptyset$, then for any $S$, $R \otimes S = S \otimes R = \emptyset$. This property maintains the upper-triangularity
of the transitive closure of $T$.

Figure~\ref{fig:otimes} gives the algorithm for multiplying two elements of the matrix. The algorithm is composed of two components.
The first component (step \ref{step:ab} in Figure~\ref{fig:otimes}) adds nonterminals, for example, $A$, to cell $(i,j)$, if there
is some $B$ and $C$ in $(i,k)$ and $(k,j)$, respectively, such that there exists a rule $A \rightarrow B\, C$ {\em and} the span
endpoints denoted by $k$ are the points where the rule specifies that
spans of $B$ and $C$ should meet.

In order to make this first component valid, we have to make sure that $k$ can indeed serve as a concatenation point for $(i,j)$. 
Step \ref{step:ab} verifies this using the concept
of configurations defined above.
To apply a rule $r : A[\alpha] \rightarrow B[\beta] \, C[\gamma]$,
we must have an entry for $(B, i, k)$ in cell $(i, k)$, 
where $i$ is a set of indices corresponding to the endpoints of $B$
selected by $\configtwo(r)$ and $k$ is a set of indices corresponding to the endpoints of $B$
selected by $[2\varphi(B)] \setminus \configtwo(r)$.  This condition is
enforced by step \ref{step:ab}\ref{step:bconfig} of Figure~\ref{fig:otimes}.
Similarly,  we must have an entry for $(C, k, j)$ in cell $(k, j)$, 
where $k$ is a set of indices corresponding to the endpoints of $C$
selected by $\configthree(r)$ and $j$ is a set of indices corresponding to the endpoints of $C$
selected by $[2\varphi(C)] \setminus \configthree(r)$.  This  is
enforced by step \ref{step:ab}\ref{step:cconfig}.
Finally, the spans defined by $B$ and $C$ must not overlap in
the string.  To guarantee that the spans do not overlap, 
we sort the endpoints of $A$ and check that each position in 
the sorted list is derived from either $B$ or $C$ as required
by the configuration of $A$ in $r$.  This check is performed
in step \ref{step:ab}\ref{step:aconfig} of Figure~\ref{fig:otimes}.

Given that $T$ is initialized to be upper-triangular, the properties
of matrix multiplication guarantee that all matrix powers of $T$ are
upper-triangular.  We now proceed to show that 
upper-triangular matrices are sufficient in terms of the grammar.
In particular, we need to show the following lemma:

\begin{lemma}\label{lem:abc}%
For each application of a single-initial rule
$A \rightarrow B \, C$, it is possible 
to create an entry for $A$ by multiplying
two upper-triangular matrices $T_1$ and $T_2$, where $T_1$ contains an entry for $B$,
and $T_2$ contains an entry for $C$. 
\end{lemma}
\begin{proof*}
A nonterminal $B$ appears in a cell above the diagonal if
its row address is smaller than its column address, which
in turn occurs if the leftmost endpoint of $B$ appears
in the row address rather than the column address.
The row address for $B$ contains the endpoints of $B$
that are also endpoints of $A$.
Our normal form for LCFRS rules ensures that
the leftmost endpoint of $B$ forms the leftmost endpoint of $A$.
Therefore the leftmost endpoint of $B$ is in $B$'s 
row address, and $B$ is above the diagonal.

The row address of nonterminal $C$ in $T_2$ must contain
the endpoints of $C$ that combine with endpoints of $B$.
For single-initial rules, these endpoints include the
leftmost endpoint of $C$, guaranteeing that $C$ appears
above the diagonal.

Because each instance of $A$ can be produced 
by combining elements of $T_1$ and $T_2$ that are above the diagonal,
each instance of $A$ can be produced by multiplying two upper-triangular matrices.
\end{proof*}

\subsubsection{Copy Operations}
The first component of the algorithm is sound, but not complete. 
If we were to use just this component in the algorithm,
then we would get in each cell $(i,j)$ of the transitive closure of $T$ a {\em subset} of the possible nonterminals that can span $m(i,j)$. The reason this happens
is that our addressing scheme is ``over-complete.'' This means that any pair of addresses $(i,j)$ and $(k,\ell)$ are equivalent if $m(i,j) = m(k,\ell)$.

We need to ensure that the transitive closure, using $\otimes$, propagates, or copies, nonterminals from one cell to its equivalents. This is done
by the second component of the algorithm, in steps 3--6. The algorithm does this kind of copying by using a set of six special ``copy'' symbols, 
$\{ \fcol, \frow, \tcol, \trow, \ucol, \urow \}$.
These symbols copy nonterminals from one cell to the other in multiple stages. 

Suppose that we need to copy a nonterminal from cell $(i,j)$ to cell $(k,\ell)$,
where $\merge(i,j) = \merge(k,\ell)$, indicating that the two cells describe the 
same set of indices in the input string.
We must move the indices in $i\cap \ell$ from the row address to the column
address, and we must move the indices in $j \cap k$ from the column address
to the row address.
We will move one index at a time, adding nonterminals to intermediate cells along the way.

\begin{figure}
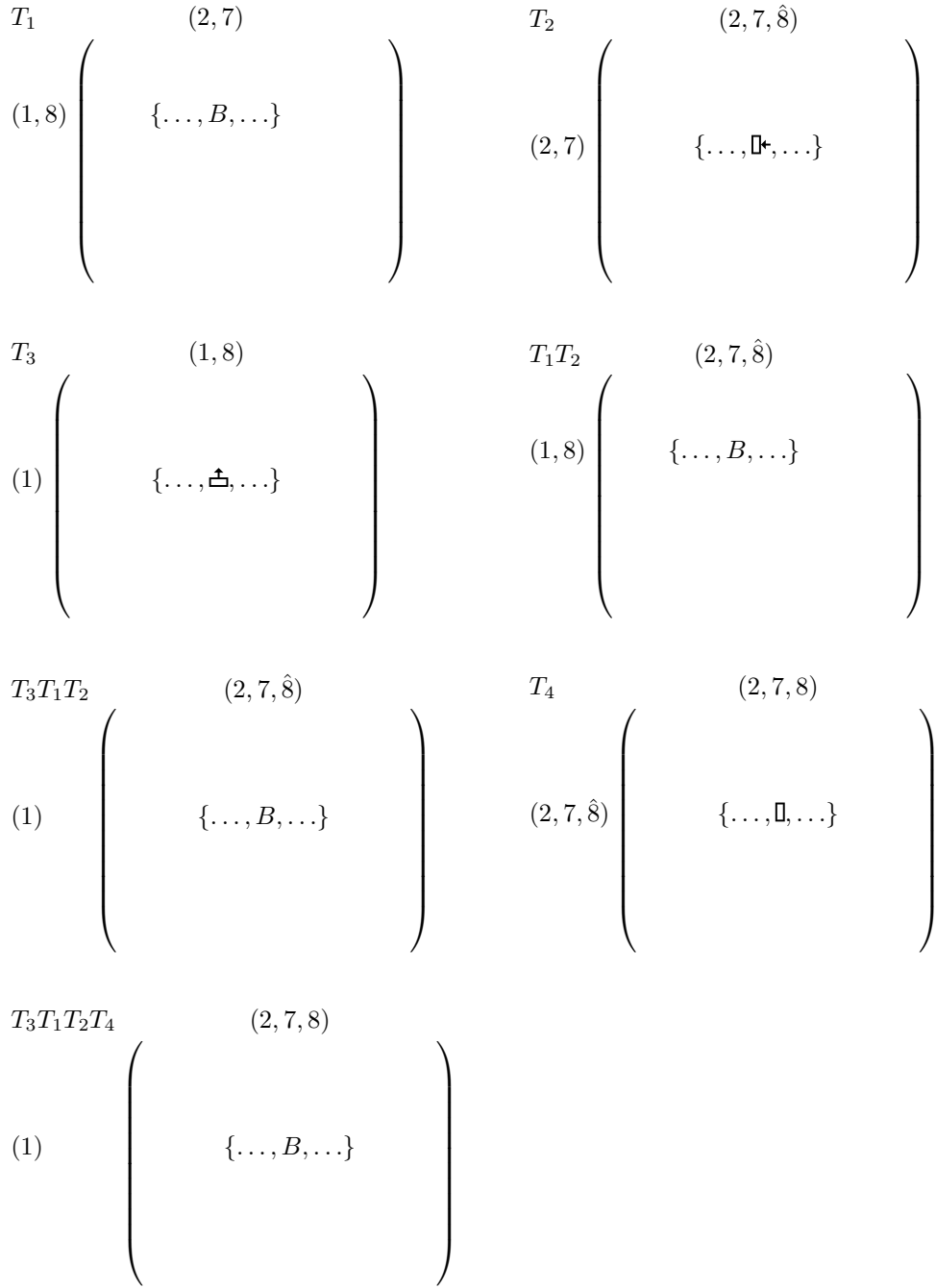


\begin{tabular}{ll}
\begin{minipage}{0.5\textwidth}
$
\bordermatrix{T_1 & &  &  (2,7) & & & \cr
                 &   &  &  &  &   &  &  \cr
                &   &  &  &    &   & &  \cr
                (1,8) &  &   &  \{ \ldots, B, \ldots \} &  &  &   &  \cr
                 &  &   &  &  &    &   &   \cr
                 &   & &  &  &  &    &   \cr
                &   &  &  & &  &   &  \cr
                &  &  & & & &    &  \cr
                 &   &  &  &  &  &  & }
$
\end{minipage}

&

\begin{minipage}{0.5\textwidth}
$
\bordermatrix{T_2 &  &  &   & (2,7,\hat 8) & & \cr
                 &   &  &  &  &   &  &  \cr
                &   &  &  &    &   & &  \cr
                 &  &   &   &  &  &   &  \cr
                (2,7) &  &   &  & \{ \ldots, \tcol, \ldots \} &    &   &   \cr
                 &   & &  &  &  &    &   \cr
                &   &  &  & &  &   &  \cr
                &  &  & & & &    &  \cr
                 &   &  &  &  &  &  & }
$
\end{minipage}
\\[7em]
\begin{minipage}{0.5\textwidth}
$
\bordermatrix{T_3 &  &  &   & (1, 8) & & \cr
                 &   &  &  &  &   &  &  \cr
                &   &  &  &    &   & &  \cr
                 &  &   &   &  &  &   &  \cr
                (1) &  &   &  & \{ \ldots, \frow, \ldots \} &    &   &   \cr
                 &   & &  &  &  &    &   \cr
                &   &  &  & &  &   &  \cr
                &  &  & & & &    &  \cr
                 &   &  &  &  &  &  & }
$
\end{minipage}
&
\begin{minipage}{0.5\textwidth}
$
\bordermatrix{T_1T_2 & &  &  (2,7,\hat 8) & & & \cr
                 &   &  &  &  &   &  &  \cr
                &   &  &  &    &   & &  \cr
                (1,8) &  &   &  \{ \ldots, B, \ldots \} &  &  &   &  \cr
                 &  &   &  &  &    &   &   \cr
                 &   & &  &  &  &    &   \cr
                &   &  &  & &  &   &  \cr
                &  &  & & & &    &  \cr
                 &   &  &  &  &  &  & }
$

\end{minipage}
\\[7em]
\begin{minipage}{0.5\textwidth}
$
\bordermatrix{T_3T_1T_2 &  &  &   & (2,7,\hat 8) & & \cr
                 &   &  &  &  &   &  &  \cr
                &   &  &  &    &   & &  \cr
                 &  &   &   &  &  &   &  \cr
                (1) &  &   &  & \{ \ldots, B, \ldots \} &    &   &   \cr
                 &   & &  &  &  &    &   \cr
                &   &  &  & &  &   &  \cr
                &  &  & & & &    &  \cr
                 &   &  &  &  &  &  & }
$
\end{minipage}
&
\begin{minipage}{0.5\textwidth}
$
\bordermatrix{T_4 &  &  &   & (2,7, 8) & & \cr
                 &   &  &  &  &   &  &  \cr
                &   &  &  &    &   & &  \cr
                 &  &   &   &  &  &   &  \cr
               (2,7,\hat 8) &  &   &  & \{ \ldots, \ucol, \ldots \} &    &   &   \cr
                 &   & &  &  &  &    &   \cr
                &   &  &  & &  &   &  \cr
                &  &  & & & &    &  \cr
                 &   &  &  &  &  &  & }
$
\end{minipage}
\\[7em]
\begin{minipage}{0.5\textwidth}
$
\bordermatrix{T_3T_1T_2T_4 &  &  &   & (2,7, 8) & & \cr
                 &   &  &  &  &   &  &  \cr
                &   &  &  &    &   & &  \cr
                 &  &   &   &  &  &   &  \cr
               (1) &  &   &  & \{ \ldots, B, \ldots \} &    &   &   \cr
                 &   & &  &  &  &    &   \cr
                &   &  &  & &  &   &  \cr
                &  &  & & & &    &  \cr
                 &   &  &  &  &  &  & }
$
\end{minipage}

\end{tabular}

\caption{An example of moving an index from the row address to the column
address.
Nonterminal $B$ in $T_1$ is copied from cell $(1,8),(2,7)$ to cell
$(1),(2,7,8)$ through three matrix multiplications.  First, multiplying
by $T_2$ on the right yields $T_1T_2$, shown in the right of the second 
row.  Multiplying this matrix by $T_3$ on the left yields $T_1T_2T_3$.
Finally, multiplying this matrix by $T_4$ on the right yields $T_1T_2T_3T_4$,
shown in the bottom row. }\label{fig:tocol}
\end{figure}

We now illustrate how our operations move a single index from a
row address to a column address (moving from column to row is similar).
Let $x$ indicate the index we wish to move, meaning that we wish to copy
a nonterminal in cell $(i,j)$ to cell $(\remove(i,x), \insertt(j,x))$.
Because we want our overall parsing algorithm to take advantage of
fast matrix multiplication, we accomplish the copy operations through a sequence
of three matrix multiplications, as shown in Figure~\ref{fig:tocol}.
The first multiplication involves 
the nonterminal $A$ in cell $(i,j)$ in the left matrix, 
and a $\tcol$ symbol in cell $(j, \insertt(j,\hat x))$
in the right matrix, resulting in a matrix with nonterminal $A$ in cell $(i, \insertt(j,\hat x))$.
This intermediate result is redundant in the sense that index $x$ appears in 
the row and index $\hat x$ appears in the column address.  To remove $x$ from the row address,
we multiply on the left with a matrix containing the symbol $\frow$
in cell $(\remove(i,x), i)$, resulting in a matrix with nonterminal $A$ in 
cell $(\remove(i,x), \insertt(j,\hat x))$.  Finally, we multiply by a third
matrix to replace the marked index $\hat x$ with the unmarked index $x$.
This is done by multiplying on the right with a matrix containing
the symbol $\ucol$ in cell $(\insertt(j,\hat x),\insertt(j,x))$.

The key idea behind the above three-step process is to copy elements from one cell
to another through intermediate cells. In matrix multiplication, only cells that share
a row or a column index actually interact when doing multiplication. Therefore,
in order to copy a nonterminal from $(i,j)$ to another cell which represents
the same set of spans, we have to copy it through cells such as $(i, \insertt(j,x))$
that share the index $i$ with $(i,j)$.

In order to guarantee that our operations copy nonterminals only into 
cells with equivalent addresses, the seed matrix contains the special symbol 
$\tcol$ only in cells $(j,k)$ such that $k=\insertt(j,\hat x)$ for some $x$.
When $\tcol$ in cell $(j,k)$ combines
with a nonterminal $A$ in cell $(i,j)$, the result contains $A$ only if
$x \in i$, guaranteeing that the index added to the column address
was originally present in the row address.
In addition, the condition that $i$ contains only unmarked indices
(in the multiplication operator) and that the condition $j$ contains only 
unmarked indices (in the initialization of the seed matrix) guarantee
that only one index is marked in the address of any non-empty matrix cell.

Similar conditions apply to the $\frow$ operation.  The seed matrix contains
$\frow$ only in cells $(i,k)$ such that $i=\remove(k,x)$ for some $x$,
guaranteeing that the operation only removes one index at a time.
Furthermore, when $\frow$ in cell $(i,k)$ combines with a nonterminal $A$ in cell
$(k,j)$, the result contains $A$ only if $\hat x \in j$.
This guarantees that the new entry includes all the original indices, meaning that 
any index we remove from the row address is still present as a marked index in the column address.

The $\ucol$ operator removes the mark on index $\hat x$ in the column 
address, completing the entire copying process.  The condition 
$|i\cup j|=\varphi(A)$ ensures that the removal of the mark from $\hat x$ does
not take place until after $x$ has been removed from the row address.

Taken together, these conditions ensure that after a sequence
of one $\tcol$, one $\frow$, and one $\ucol$, $A$ is copied into
all cells having the form $(\remove(i,x), \insertt(j, x))$ for some $x$.

To move an index from the column address to the row address, we use
one $\trow$ operation followed by one $\fcol$ operation and one $\urow$ operation.  The
conditions on these three special symbols are analogous to the conditions on
$\tcol$, $\frow$, and $\ucol$ outlined above, and ensure that we copy from cell 
$(i,j)$ to cells of the form $(\insertt(i,x), \remove(j,x))$ for some $x$.

We now show that matrix powers of the upper-triangular seed matrix $T$
copy nonterminals between all equivalent cells above the diagonal.

\begin{lemma}\label{lem:1copy}%
Let $(i,j)$ and $(k,\ell)$ be unmarked matrix addresses, 
in a seed matrix $T$ indexed by
row and column addresses from $N(d)$ where
$d > \min\{|i|,|j|\}$ and $d > \min\{|k|, |\ell|\}$.  Assume that 
$\min i = \min k$ and either 
$k = \remove(i,x)$ and $\ell = \insertt(j,x)$ for some $x$, or 
$k = \insertt(i,x)$ and $\ell = \remove(j,x)$ for some $x$.
If $A$ appears in cell $(i,j)$ of $T^{(n)}$, then
$A$ appears in cell $(k, \ell)$ of $T^{(n+3)}$.
Furthermore, the copy operations do not introduce 
nonterminals into any other cells with unmarked addresses.
\end{lemma}
\begin{proof*}
The condition on $d$ guarantees that we can form
row and column addresses long enough to hold the redundant 
representations with one address shared between row and column. 
This condition is only relevant in the case where $i$, $j$, $k$, and $\ell$
are all of the same length; in this case we need to construct
temporary indices with length one greater, 
as in the example in Figure~\ref{fig:tocol}.

$A$ can be added to cell $(k,\ell)$ through a sequence 
of three matrix multiplications by combining with 
symbols $\tcol$, $\frow$, and $\ucol$ or with 
$\trow$, $\fcol$, and $\urow$.  Because $T^{(n)}$ is 
upper triangular, $\min i = \min i\cup j$, meaning that
$A$'s leftmost index is in its row address.  The condition
$\min i = \min k$ implies that we are not moving this 
leftmost index from row to column.  The addresses
of the three copy symbols required are all formed 
by adding or removing $x$ or $\hat x$ to the row and
column addresses $(i,j)$; because the leftmost index of $i$
is not modified, the copy symbols that are required are all
above the diagonal, and are present in the seed matrix $T$.
Therefore, $A$ appears in cell $(k,\ell)$ of $T^{(n+3)}$.

To see that nonterminals are not introduced into any other cells,
observe that $\urow$ and $\ucol$ are the only symbols that
introduce nonterminals into unmarked addresses.  They can
only apply when a marked index is present, and when
the total number indices is $2\varphi(A)$.  This can only occur
after either $\tcol$ has introduced a marked index and 
$\frow$ removed the corresponding unmarked index, or
$\trow$ has introduced a marked index and 
$\fcol$ removed the corresponding unmarked index
\end{proof*}

Putting together sequences of these operations to move indices,
we get the following lemma:

\begin{lemma}\label{lemma:copy}%
Let $(i,j)$ and $(k,\ell)$ be matrix addresses such that 
$\merge(i,j) = \merge(k,\ell)$, in a seed matrix $T$ indexed by
row and column addresses from $N(d)$ where
$d > \min\{|i|,|j|\}$ and $d > \min\{|k|, |\ell|\}$.
Then, for any nonterminal $A$ in cell $(i,j)$ in $T^{(n)}$,
$A$ will also appear in cell $(k,\ell)$ of the power matrix $T^{(n+6d)}$.
\end{lemma}
\begin{proof*}
Nonterminal $A$ can be copied through a series of intermediate 
cells by moving one index at a time from $i$ to $\ell$, and 
from $j$ to $k$.  We begin by moving indices from either
the row address $i$ to the column address if $|i| > |j|$,
or from the column address $j$ to the row address otherwise.
We must move up to $d$ indices from row to column, and $d$ 
indices from column to row.  Each move takes three matrix 
multiplications, for a total of $6d$ matrix multiplications.
\end{proof*}



\subsection{Determining the Contact Rank}
\label{sec:d}

The dimensions of the matrix $T$ (and its transitive closure) are $|N| \times |N|$. The set $N$ is of size $O(n^d)$, where $d$ is a function of the grammar.
When a given pair of cells in two matrices of the type of $T$ are multiplied, we are essentially combining endpoints from the first multiplicand column address
with endpoints from the second multiplicand row address. 
As such, we have to ensure that $d$ allows us to generate all possible sequences of endpoints that could
potentially combine with a given fixed LCFRS.

We refer to the endpoints at which a rule's r.h.s.\ nonterminals meet
as {\bf combining points}.  For example, in the simple case of a CFG with a rule $\mathrm{S} \rightarrow \mathrm{NP}\,\mathrm{VP}$,
there is one combining point where $\mathrm{NP}$ and $\mathrm{VP}$ meet.  
For the TAG rule shown in Figure~\ref{fig:tag},
there are two combining points where nonterminals $B$ and $C$ meet. 
For each rule $r$ in the
LCFRS grammar, we must be able to access the combining points as row
and column addresses in order to apply the rule with matrix multiplication.
Thus, $d$ must be at least the maximum number of combining points of any rule
in the grammar.  The number of combining points $\delta(r)$ for a rule $r$
can be computed by comparing the number of spans on the l.h.s.\ and 
r.h.s.\ of the rule:
\begin{equation}
\delta(A[\alpha] \rightarrow B[\beta] \, C[\gamma]) = \varphi(C) + \varphi(B) - \varphi(A). \label{eq:d1}
\end{equation}

Note that $\delta(r)$ depends only on the skeleton of $r$ (see \S\ref{section:notation}), and therefore
it can be denoted by $\delta(A \rightarrow B\, C)$.\footnote{To see that Eq.~\refeq{eq:d1} is true, consider that if we take $\varphi(B) + \varphi(C)$ variables
from the spans of the r.h.s.  and try to combine them together to $\varphi(A)$ sequences per span of the l.h.s., we will get $\varphi(B) + \varphi(C) - \varphi(A)$
points where variables ``touch.'' If $\varphi(A) = 1$, then this is clearly true. For $\varphi(A) > 1$, consider that for each span, we ``lose'' one contact point.}

For each nonterminal on the r.h.s.\ of the rule, the address of 
its matrix cell consists of the combination points in one dimension
(either row or column), and the other points in the other dimension
of the matrix.  For r.h.s.\ nonterminal $B$ in rule 
$A \rightarrow B \, C$,
the number of non-combination endpoints is:
\begin{equation}
2\varphi(B) - \delta(A \rightarrow B \, C).
\end{equation}

Thus, taking the maximum size over all addresses in the grammar,
the largest addresses needed are of length:
\begin{equation}
d = \max_{A \rightarrow B \, C \in \rulessym} \max \left\{ 
\begin{array}{c}
\delta(A \rightarrow B \, C), \\
2\varphi(B) - \delta(A \rightarrow B \, C),\\
2\varphi(C) - \delta(A \rightarrow B \, C)
\end{array} \right\}.
\end{equation}

We call this number the {\bf contact rank} of the grammar. 
As examples, the contact rank of a CFG is one, 
while the contact rank of a TAG is two.
A simple algebraic manipulation shows that the contact rank can be expressed as follows:
\begin{equation}
d = \max_{A \rightarrow B \, C \in \rulessym} \max \left\{
\begin{array}{c}
\varphi(A) + \varphi(B) - \varphi(C), \\ \varphi(A) - \varphi(B) + \varphi(C), \\ -\varphi(A) + \varphi(B) + \varphi(C)
\end{array} \right\}.
\end{equation}

We require our grammars to be in single-initial form, 
as described in \S\ref{section:lcfrs}.
Because the process of converting an LCFRS grammar to single-initial
form increases its fan-out by at most one, the contact rank is also 
increased by at most one.

\subsection{Balanced Grammars}
\label{section:balanced}

We define the {\bf configuration set} of a nonterminal $A$ to the
the set of all configurations (\S\ref{sec:config})
in which $A$ appears in a grammar rule,
including both appearances in the r.h.s.\ and as the l.h.s.
\[
\config(a) = \left( \bigcup_{r:\mathrm{lhs}(r) = A} \{\configone(r)\} \right) \cup 
       \left( \bigcup_{r:\mathrm{rhs1}(r) = A} \{\configtwo(r)\} \right) \cup 
       \left( \bigcup_{r:\mathrm{rhs2}(r) = A} \{\configthree(r)\} \right)
\]
For example, in a CFG, the configuration set of any 
nonterminal is $\{ \{ 1 \} \}$, because, as shown in 
\S\ref{sec:config}, nonterminals are always used in 
the unique configuration $\{ 1 \}$.
For TAG, the configuration set of any nonterminal is $\{ \{ 1, 4 \} \}$
because, as in CFG, nonterminals are always used in the same configuration.

A configuration $\configsym$ of nonterminal $B$ is {\bf balanced} if $|\configsym| = \varphi(B)$.
This means that the number of contact points and non-contact points are the same.

The contact rank $d$ defined in the previous section is the maximum size
of any configuration of any nonterminal in any rule.
For a given nonterminal $B$, if $\varphi(B)<d$, then we can copy 
entries between equivalent cells.  To see this, suppose that
we are moving from cell $(i,j)$ to $(k, \ell)$ where the length of $i$ 
is greater than the length of $j$.  As long as we move the first 
index from row to column, rather than from column to row, the intermediate
results will require addresses no longer than the length of $i$.

However, if $\varphi(B)=d$, then
every configuration in which $B$ appears is balanced:
\[
\forall \configsym \in \config(B) \quad |\configsym| = \varphi(B)
\]
If $\varphi(B)=d$ and $B$ appears 
in more than one configuration, that is, $|\config(B)| > 1$,
it is impossible to copy 
entries for $B$ between the cells using a matrix of size $(2n)^d$.
This is because we cannot move
indices from row to column or from column to row without creating an
intermediate row or column address of length greater than $d$
as a result of the first $\tcol$ or $\trow$ operation.

We define a {\bf balanced grammar} to be a grammar containing a 
nonterminal $B$ such that $\varphi(B)=d$, and $|\config(B)| > 1$.
As examples, a CFG is not balanced because, while, for each
nonterminal $B$, $\varphi(B)=d=1$,
the number of configurations $|\config(B)|$ is one.
Similarly, TAG is not balanced, because each nonterminal has
only one configuration.  Inversion Transduction Grammars (ITGs)
are balanced, because, for each nonterminal $B$, $\varphi(B)=d=2$,
and nonterminals can be used in two configurations,
corresponding to straight and inverted rules. 

The following condition will determine which of 
two alternative methods we use for the top level of 
our parsing algorithm.

\begin{cond}
{\bf Unbalanced Grammar Condition} There is no nonterminal $B$ such that $\varphi(B) = d$ and 
$|\config(B)| > 1$.
\label{cond:abg}
\end{cond}

This condition guarantees that we can move nonterminals as 
necessary with matrix multiplication:

\begin{lemma}\label{lemma:balanced}%
Let $(i,j)$ and $(k,\ell)$ be matrix addresses such that 
$\merge(i,j) = \merge(k,\ell)$.
Under Condition~\ref{cond:abg}, for any nonterminal $A$ in cell $(i,j)$ in $T^{(n)}$,
$A$ will also appear in cell $(k,\ell)$ of the power matrix $T^{(n+6d)}$.
\end{lemma}
\begin{proof*}
The number of $A$'s endpoints is $2\varphi(A) = |i|+|j| = 
|k| + |\ell|$.
If the grammar is not balanced, then 
$d > \varphi(A)$, and therefore $d > \min\{|i|,|j|\}$ and 
$d > \min\{|k|, |\ell|\}$.
By Lemma~\ref{lemma:copy}, $A$ will appear in cell $(k,\ell)$ of the power matrix $T^{(n+6d)}$.
\end{proof*}

\subsection{Computing the Transitive Closure of $T$}
\label{section:transitive-closure}

The transitive closure $T^+$ of a matrix $T$ is the result
of repeated applications of the matrix multiplication operator described in Eq.~\refeq{eq:a}. With $T$ being the seed matrix, we define
\begin{equation}
T^+ = T^{(1)} \cup T^{(2)} \cup \cdots, 
\end{equation}
 where $T^{(i)}$ is defined recursively as:
\begin{align}
T^{(1)} & = T \\
T^{(i)} & = \bigcup_{j=1}^{i-1} \left( T^{(j)} \otimes T^{(i-j)} \right).
\end{align}

Under Condition~\ref{cond:abg}, one can show that given an LCFRS derivation tree $t$ over the input
string, each node in $t$ must appear in the transitive 
closure matrix $T^+$.  Specifically, for each node
in $t$ representing nonterminal $A$ spanning
endpoints $\{ (\ell_1,\ell_2), (\ell_3,\ell_4), \ldots, (\ell_{2\varphi(A)-1},\ell_{2\varphi(A)}) \}$, at each cell $T^+_{i, j}$ in the matrix 
such that $m(i,j) = \{ (\ell_1,\ell_2), (\ell_3,\ell_4), \ldots, (\ell_{2\varphi(A)-1},\ell_{2\varphi(A)}) \}$, contains 
$A$.   This leads to the following result:

\begin{lemma}\label{lemma:chart}%
Under Condition~\ref{cond:abg}, the transitive closure of $T$ is such that $[T^+]_{ij}$ represents the set of nonterminals that are derivable for the given spans in $m(i,j)$.
\end{lemma}
\begin{proof*}
The proof is by induction over 
the length of the LCFRS derivations.  By Lemma~\ref{lem:abc}, derivations
consisting of a single rule $A[\alpha] \rightarrow B[\beta] \, C[\gamma]$
produce $A \in T^{(2)}$ for $i$ and $j$ 
corresponding the non-combination points of $B$ and $C$.
For all other $i$ and $j$ such that $m(i,j) = \{ (\ell_1,\ell_2), (\ell_3,\ell_4), \ldots, (\ell_{2\varphi(A)-1},\ell_{2\varphi(A)}) \}$,
an entry is produced in $T^{(6d)}_{ij}$ 
by Lemma~\ref{lemma:balanced}.
By induction, $T^{s(6d+2)}$ contains entries for all
LCFRS derivations of depth $s$, and $T^+$ contains 
entries for all LCFRS derivations of any length. 

In the other direction, we need to show that all entries
$A$ in $T^+$ correspond to a valid LCFRS derivation
of nonterminal $A$ spanning endpoints $m(i,j)$.
This can be shown by induction over the number of 
matrix multiplications.  During each multiplication,
entries created in the product matrix correspond either
to the application of an LCFRS rule with l.h.s.\ $A$, or to the movement
of an index between row and column address for a previously
recognized instance of $A$.
\end{proof*}

The transitive closure still yields a useful result, even when Condition~\ref{cond:abg} does not hold.
To show how it is useful, we need to define the ``copying'' operator, $\Pi$, which takes a matrix $T'$ of the same type of $T$, and sets $\Pi(T')$ using the following procedure:

\begin{enumerate}

\item Define $e(i,j) = \{ (i', j') \mid m(i',j') = m(i,j) \}$, i.e. the set of equivalent configurations to $(i,j)$.

\item Set $[\Pi(T')]_{ij} = \displaystyle\bigcup_{(i',j') \in e(i,j)} A_{i'j'} $.

\end{enumerate}

This means that $\Pi$ takes a completion step, and copies all nonterminals between all equivalent addresses in $T'$.
Note that the $\Pi$ operator can be implemented such that it operates in time $O(n^d)$. All it requires is just taking $O(n^d)$ unions of sets (corresponding to the sets of nonterminals
in the matrix cells), where each set is of size $O(1)$ with respect to the sentence length 
(i.e. the size is only a function of the grammar).

This procedure leads to a recognition algorithm for binary LCFRS that do not satisfy Condition~\ref{cond:abg}
(we also assume that these binary LCFRS would not have unary cycles or $\epsilon$ rules).
This algorithm is given in Figure~\ref{fig:parse-alg2}. It operates by iterating through transitive closure steps and copying steps until convergence.
When we take the transitive closure of $T$, we are essentially computing a subset of the derivable nonterminals. Then, the copying step (with $\Pi)$ propagates nonterminals through
equivalent cells. Now, if we take the transitive closure again, and there is any way to derive new nonterminals because of the copying step, the resulting matrix will have at least
one new nonterminal. Otherwise, it will not change, and as such, we recognized all possible derivable nonterminals in each cell.

\begin{lemma}
For any single-initial LCFRS,
when step 2 of the algorithm in Figure~\ref{fig:parse-alg2}
converges, $T$ is such that $[T]_{ij}$ represents the set of nonterminals that are derivable for the given spans in $m(i,j)$.
\label{lemma:b}
\end{lemma}
\begin{proof*}
Any LCFRS derivation of a nonterminal can be decomposed into a sequence of 
rule applications and copy operations, and by induction over
the length of the derivation, all derivations will be found.
Each matrix operation only produces derivable LCFRS nonterminals,
and by induction over the number of steps of the algorithm,
only derivable nonterminals will be found.
\end{proof*}

\begin{figure}

\begin{tabular}{ll}
\begin{minipage}{0.5\textwidth}
$
\bordermatrix{G_B & &  &  (2,7) & & & \cr
                 &   &  &  &  &   &  &  \cr
                &   &  &  &    &   & &  \cr
                (1,8) &  &   &  1 &  &  &   &  \cr
                 &  &   &  &  &    &   &   \cr
                 &   & &  &  &  &    &   \cr
                &   &  &  & &  &   &  \cr
                &  &  & & & &    &  \cr
                 &   &  &  &  &  &  & }
$
\end{minipage}

&

\begin{minipage}{0.5\textwidth}
$
\bordermatrix{H_C &  &  &   & (4,5) & & \cr
                 &   &  &  &  &   &  &  \cr
                &   &  &  &    &   & &  \cr
                 &  &   &   &  &  &   &  \cr
                (2,7) &  &   &  & 1 &    &   &   \cr
                 &   & &  &  &  &    &   \cr
                &   &  &  & &  &   &  \cr
                &  &  & & & &    &  \cr
                 &   &  &  &  &  &  & }
$
\end{minipage}
\\[7em]
\begin{minipage}{0.5\textwidth}
$
\bordermatrix{I_{BC} = G_BH_C &  &  &   & (4,5) & & \cr
                 &   &  &  &  &   &  &  \cr
                &   &  &  &    &   & &  \cr
                 &  &   &   &  &  &   &  \cr
                (1,8) &  &   &  & 1 &    &   &   \cr
                 &   & &  &  &  &    &   \cr
                &   &  &  & &  &   &  \cr
                &  &  & & & &    &  \cr
                 &   &  &  &  &  &  & }
$
\end{minipage}
&

\end{tabular}
\caption{Reduction of transitive closure to Boolean matrix multiplication. 
Boolean matrix operations implementing the matrix
multiplication example of Section~\ref{section:sketch}. }\label{fig:Boolean}
\end{figure}

\subsubsection{Reduction of Transitive Closure to Boolean Matrix Multiplication}
Valiant showed that his algorithm for computing the multiplication of two matrices, in terms of
a multiplication operator similar to ours, can be reduced to the problem of Boolean matrix multiplication.
His transitive closure algorithm requires as a black box this two-matrix multiplication algorithm.

We follow here a similar argument. We can use Valiant's algorithm for the computation of the transitive closure,
since our multiplication operator is distributive (with respect to $\cup$). To complete our argument, we need
to show, similarly to Valiant, that the product of two matrices using our multiplication operator can be reduced
to Boolean matrix multiplication.

Consider the problem of multiplication a matrix $T_1$ and $T_2$, and say $T_1 \otimes T_2 = T_3$. To reduce it to Boolean matrix multiplication,
we create $2|\mathcal{R}|$ pairs of matrices, $G_r$ and $H_r$, where $r$ ranges over $\mathcal{R}$.
The size of $G_r$ and $H_r$ is $N \times N$. 
If $r = A[\alpha] \rightarrow B[\beta] C[\gamma]$, 
we set $[G_r]_{ik}$ to be 1 if the nonterminal $B$ appears
in $[T_1]_{ik}$ and $B$, $i$, and $k$ meet  the conditions of step \ref{step:ab}\ref{step:bconfig} of Figure~\ref{fig:otimes}.
Similarly, we set $[H_r]_{kj}$ to be 1 if the nonterminal $C$ appears in $[T_2]_{kj}$ and $C$, $k$, and $j$ meet the conditions of step \ref{step:ab}\ref{step:cconfig}.
All other cells, in both $G_r$ and $H_r$, are set to 0. Note that $G_r$ and $H_r$ for all $r \in \mathcal{R}$ are upper
triangular Boolean matrices.

In addition, we create $2|\nts|$ pairs of matrices, $G_A$ and $H_A$, where $A$ ranges over the set of nonterminals $\nts$.
We set $[G_A]_{ik}$ to be 1 if the nonterminal $A$  appears
in $[T_1]_{ik}$, regardless the conditions of step \ref{step:ab}\ref{step:bconfig} of Figure~\ref{fig:otimes}.
Similarly, we set $[H_A]_{kj}$ to be 1 if the nonterminal $A$ appears in $[T_2]_{kj}$, 
regardless of the conditions of step \ref{step:ab}\ref{step:cconfig}.
All other cells, in both $G_A$ and $H_A$, are set to 0. 
Again, $G_A$ and $H_A$ for all $A \in \nts$ are upper
triangular Boolean matrices.

Finally, we create six additional matrices, for each element in the set $\{ \fcol, \frow, \tcol, \trow, \ucol, \urow \}$.
These matrices indicate the positions in which each symbol
appears in the seed matrix $T$ defined in Figure~\ref{fig:seed-matrix}:
\begin{enumerate}
\item $G_{\frow}$, for which $[G_{\frow}]_{ij} = 1$ only if $(\frow, i, j) \in T$. 
\item $H_{\tcol}$, for which $[H_{\tcol}]_{ij} = 1$ only if $(\tcol, i, j) \in T$. 
\item $H_{\ucol}$, for which $[H_{\ucol}]_{ij} = 1$ only if $(\ucol, i, j) \in T$. 
\item $G_{\trow}$, for which $[G_{\trow}]_{ij} = 1$ only if $(\trow, i, j) \in T$. 
\item $H_{\fcol}$, for which $[H_{\fcol}]_{ij} = 1$ only if $(\fcol, i, j) \in T$. 
\item $G_{\urow}$, for which $[G_{\urow}]_{ij} = 1$ only if $(\urow, i, j) \in T$. 
\end{enumerate}

Now, for each rule $r \in \mathcal{R}$, we compute the matrix $I_{r} = G_r H_r$. The total number of matrix multiplications
required is $|\mathcal{R}|$, which is constant in $n$. Now, $T_3$ can be obtained by
multiplying these matrices, and applying the conditions of Figure~\ref{fig:otimes}:

\begin{enumerate}

\item For each $A \in \nts$, for each rule $r = A \rightarrow B\, C$, check whether $[I_{r}]_{ij} = 1$. If step \ref{step:ab}\ref{step:aconfig} is
satisfied for $A$, $i$, and $j$, then add $(A,i,j)$ to $[T_3]_{ij}$.

\item For each $A \in \nts$, compute $J_A = G_A H_{\tcol}$. For each $(i,j)$, add $A$ to $[T_3]_{ij}$ 
if $\hat x \in j$ and $x \in i$ for some $x$,
and $[J_A]_{ij} = 1$.

\item For each $A \in \nts$, compute $J_A = G_{\frow} H_A$. For each $(i,j)$, add $A$ to $[T_3]_{ij}$ 
$\hat x \in j$ and $x \not\in i$ and for some $x$,
and $[J_A]_{ij} = 1$.

\item For each $A \in \nts$, compute $J_A = G_A H_{\ucol}$. For each $(i,j)$, add $A$ to $[T_3]_{ij}$ 
if $|i \cup j| = 2\varphi(A)$,
and $[J_A]_{ij} = 1$.

\item For each $A \in \nts$, compute $J_A = G_{\trow} H_A$. For each $(i,j)$, add $A$ to $[T_3]_{ij}$ 
if $\hat x \in i$ and $x \in j$ for some $x$,
and $[J_A]_{ij} = 1$.

\item For each $A \in \nts$, compute $J_A = G_A H_{\fcol}$. For each $(i,j)$, add $A$ to $[T_3]_{ij}$ 
$\hat x \in i$ and $x \not\in j$ for some $x$,
and $[J_A]_{ij} = 1$.

\item For each $A \in \nts$, compute $J_A = G_{\urow} H_A$. For each $(i,j)$, add $A$ to $[T_3]_{ij}$ 
if $|i \cup j| = 2\varphi(A)$,
and $[J_A]_{ij} = 1$.

\end{enumerate}

\begin{lemma}%
\label{lemma:mult}%
The matrix product operation for two matrices of size $(2n)^d \times (2n)^d$
can be computed in time $O(n^{\omega d})$, if two $m\times m$ 
Boolean matrices can be multiplied in time $O(m^\omega)$.
\end{lemma}

\begin{proof*}
The result of the algorithm above 
is guaranteed to be the same as the result of
matrix multiplication using the $\otimes$ operation of Figure~\ref{fig:otimes}
because it considers all combinations of $i$, $j$, and $k$
and all pairs of nonterminals and copy symbols, and 
applies the same set of conditions.  This is possible
because each of the conditions in Figure~\ref{fig:otimes}
applies either to a pair $(i,k)$ or $(k,j)$, in 
which case we apply the condition to input matrices
to the Boolean matrix multiplication, or to the pair $(i,j)$,
in which case we apply the condition to the result
of the Boolean matrix multiplication.  Crucially,
no condition in Figure~\ref{fig:otimes} involves $i$, $j$, and $k$
simultaneously.

The Boolean matrix algorithm takes time
$O(n^{\omega d})$ for each matrix multiplication,
while the pre- and post-processing steps for each
matrix multiplication take only $O(n^{2d})$.
The number of Boolean matrix multiplications depends on the grammar,
but is constant with respect to $n$, yielding an
overall runtime of $O(n^{\omega d})$.  
\end{proof*}


The final parsing algorithm is given in Figure~\ref{fig:parse-alg}. It works by computing the seed matrix $T$, and then finding its transitive closure. Finally, it checks whether the
start symbol appears in a cell with an address that spans the whole string. If so, the string is in the language of the grammar.

\begin{figure}

\framebox{\parbox{\columnwidth}{

{\bf Inputs:} An LCFRS grammar as defined in \S\ref{section:lcfrs} that satisfies Condition~\ref{cond:abg} and a sentence $w_1 \cdots w_n$. \\

{\bf Outputs:} $\mathrm{True}$ if $w_1 \cdots w_n$ is in the language of the grammar, $\mathrm{False}$ otherwise. \\

{\bf Algorithm:}

\begin{enumerate}

\item Compute $T$ as the seed matrix using the algorithm in Figure~\ref{fig:seed-matrix}.

\item Compute the transitive closure of $T$ with the multiplication operator in Figure~\ref{fig:otimes} and using Boolean matrix multiplication (\S\ref{section:transitive-closure}).

\item Return $\mathrm{True}$ if $(\startsym,(0),(n))$ belongs to the cell $((0),(n))$ in the computed transitive closure, and $\mathrm{False}$ otherwise.

\end{enumerate}

}}

\caption{\label{fig:parse-alg} Algorithm for recognizing binary linear context-free rewriting systems when Condition~\ref{cond:abg} is satisfied by the LCFRS.}
\end{figure}

\section{Computational Complexity Analysis}

As mentioned in the previous section, the algorithm in Figure~\ref{fig:parse-alg}
finds the transitive closure of 
a matrix under our definition of matrix multiplication.
The operations $\cup$ and $\otimes$ used in our matrix
multiplication distribute.
The $\otimes$ 
operator takes the cross product of two sets, and applies
a filtering condition to the results; the fact that
$(x \otimes y) \cup(x \otimes z) = x\otimes(y \cup x)$
follows from the fact that it does not matter whether 
we take the cross product of the union, or the union of
the cross product.  However, unlike in the case of
standard matrix multiplication, our $\otimes$ operation
is not associative.  In general, $x \otimes (y \otimes z)
\neq (x \otimes y) \otimes z$, because the combination
of $y$ and $z$ may be allowed by the LCFRS grammar, while
the combination of $x$ and $y$ is not.

\begin{lemma}\label{lemma:closure}%
The transitive closure of a matrix of size $(2n)^d \times (2n)^d$
can be computed in time $O(n^{\omega d})$, if $2 < \omega < 3$,
and two $m\times m$ Boolean matrices can be multiplied in time $O(m^\omega)$.
\end{lemma}
\begin{proof*}
We can use the algorithm of Valiant for finding the closure
of upper triangular matrices
under distributive, non-associative matrix multiplication.
Because we can perform one matrix product in
time $O(n^{\omega d})$ by Lemma~\ref{lemma:mult}, the algorithm of \citet[Theorem 2]{valiant75}
can be used to compute transitive closure also in time $O(n^{\omega d})$.
\end{proof*}

When Valiant's paper
was published, the best well-known algorithm known for such
multiplication was Strassen's algorithm, with $M(n) = O(n^{2.8704})$.
Since then, it is known that $M(n) = O(n^{\omega})$ for $\omega < 2.38$ (see
also \S\ref{section:introduction}). There are ongoing attempts to further
reduce $\omega$, or find lower bounds for $M(n)$.

The algorithm for transitive closure gives one of the main results of this article:

\begin{theorem}
A single-initial binary LCFRS meeting Condition~\ref{cond:abg} can be parsed in time $O(n^{\omega d})$, 
where $d$ is the contact rank of the grammar,
$2 < \omega < 3$,
and two $m\times m$ Boolean matrices can be multiplied in time $O(m^\omega)$.
\end{theorem}
\begin{proof*}
By Lemma~\ref{lemma:closure}, step 2 of  the algorithm in Figure~\ref{fig:parse-alg} takes $O(n^{\omega d})$.  By Lemma~\ref{lemma:chart}, the 
result of step 2 gives all nonterminals that are derivable for the given spans in $m(i,j)$.
\end{proof*}

Parsing a binary LCFRS rule with standard chart parsing techniques
requires time $O(n^{\varphi(A)+\varphi(B)+\varphi(C)})$.
Let $p = \max_{A \rightarrow B \, C \in \rulessym} \left( \varphi(A)+\varphi(B)+\varphi(C) \right)$.
The worst-case complexity of LCFRS chart parsing techniques is $O(n^{p})$.
We can now ask the question: in which case the algorithm in Figure~\ref{fig:parse-alg} is asymptotically more efficient
than standard chart parsing techniques with respect to $n$? 
That is, in which cases is $n^{d \omega} = o(n^{p})$?

Clearly, this would hold whenever $d \omega < p$. By definition of $d$ and $p$, a sufficient
condition for that is that for any rule $A \rightarrow B \, C \in \rulessym$ it holds
that:\footnote{For two sets of real numbers, $X$ and $Y$, it holds that if for all $x \in X$ there is a $y \in Y$ such that $x < y$, then $\max X < \max Y$.}
\begin{equation}
\max \left\{ \begin{array}{c} \varphi(A) + \varphi(B) - \varphi(C), \\ \varphi(A) - \varphi(B) + \varphi(C), \\ -\varphi(A) + \varphi(B) + \varphi(C) \end{array} \right\} < \frac{1}{\omega} \left( \varphi(A) + \varphi(B) + \varphi(C) \right).
\end{equation}

This means that for any rule, the following conditions should hold:
\begin{align}
\omega (\varphi(A) + \varphi(B) - \varphi(C) ) & < \varphi(A) + \varphi(B) + \varphi(C), \\
\omega (\varphi(A) - \varphi(B) + \varphi(C) ) & < \varphi(A) + \varphi(B) + \varphi(C), \\
\omega (- \varphi(A) + \varphi(B) + \varphi(C) ) & < \varphi(A) + \varphi(B) + \varphi(C).
\end{align}

Algebraic manipulation shows that this is equivalent to having:
\begin{align}
\varphi(A) + \varphi(B) < \left( \displaystyle\frac{\omega + 1}{\omega-1}\right)\varphi(C), \\
\varphi(B) + \varphi(C) < \left( \displaystyle\frac{\omega + 1}{\omega-1}\right)\varphi(A), \\
\varphi(C) + \varphi(A) < \left( \displaystyle\frac{\omega + 1}{\omega-1}\right)\varphi(B).
\end{align}

For the best well-known algorithm for matrix multiplication, it holds that:
\begin{align}
\displaystyle\frac{\omega + 1}{\omega-1} > 2.44.
\end{align}

For Strassen's algorithm, it holds that:
\begin{align}
\displaystyle\frac{\omega + 1}{\omega-1} > 2.06.
\end{align}

\begin{figure}

\framebox{\parbox{\columnwidth}{

{\bf Inputs:} An LCFRS grammar as defined in \S\ref{section:lcfrs} and a sentence $w_1 \cdots w_n$. \\

{\bf Outputs:} $\mathrm{True}$ if $w_1 \cdots w_n$ is in the language of the grammar, $\mathrm{False}$ otherwise. \\

{\bf Algorithm:}

\begin{enumerate}

\item Compute $T$ as the seed matrix using the algorithm in Figure~\ref{fig:seed-matrix}.

\item Repeat until $T$ does not change: $T \leftarrow \left( \Pi(T) \right)^+$.

\item Return $\mathrm{True}$ if $(\startsym,(0),(n))$ belongs to the cell $((0),(n))$ in the computed transitive closure, and $\mathrm{False}$ otherwise.

\end{enumerate}

}}

\caption{\label{fig:parse-alg2} Algorithm for recognizing binary LCFRS when Condition~\ref{cond:abg} is not necessarily satisfied by the LCFRS.}
\end{figure}

We turn now to analyze the complexity of the algorithm in Figure~\ref{fig:parse-alg2}, giving the main result of this article for arbitrary LCFRS: 
\begin{theorem}
A single-initial binary LCFRS can be parsed in time $O(n^{\omega d+1})$, 
where $d$ is the contact rank of the grammar,
$2 < \omega < 3$,
and two $m\times m$ Boolean matrices can be multiplied in time $O(m^\omega)$.\end{theorem}
\begin{proof*}
The algorithm of Figure~\ref{fig:parse-alg2} works by iteratively applying the transitive closure and the
copying operator until convergence. 
At convergence, we have recognized all derivable nonterminals by
Lemma~\ref{lemma:b}.
Each transitive closure has the asymptotic complexity of $O(n^{\omega d})$ by Lemma~\ref{lemma:closure}. Each $\Pi$ application has the asymptotic complexity of $O(n^d)$.
As such, the total complexity is $O(t n^{\omega d})$, where $t$ is the number of iterations required to converge. At each iteration, we discover at least one new nonterminal. The
total number of nodes in the derivation for the recognized string is $O(n)$ (assuming no unary cycles or $\epsilon$ rules). As such $t = O(n)$, and the total complexity
of this algorithm is $O(n^{\omega d + 1})$.
\end{proof*}

\section{Applications}

Our algorithm is a recognition algorithm which is applicable to binary LCFRS.
As such, our algorithm can be applied to any LCFRS, by first
reducing it to a binary LCFRS.  We discuss results for specific classes of LCFRS
in this section, and return to the general binarization process  
in \S\ref{section:binarization-strategy}.

LCFRS subsumes context-free grammars, which was the formalism that \citet{valiant75} focused on. Valiant showed that the problem of CFG recognition
can be reduced to the problem of matrix multiplication, and as such, the complexity of CFG 
recognition in that case is $O(n^{\omega})$.
Our result generalizes Valiant's result. CFGs (in Chomsky normal form) can be reduced to a binary LCFRS with $f=1$. As such, $d=1$ for CFGs, and our algorithm yields a complexity of $O(n^{\omega})$.
(Note that CFGs satisfy Condition~\ref{cond:abg}, and therefore we can use a single transitive closure step.)

LCFRS is a broad family of grammars, and it subsumes many other well-known grammar formalisms, some 
of which were discovered or developed independently of LCFRS.
Two such formalisms are tree-adjoining grammars \cite{joshi1997tree} and synchronous context-free grammars. In the next two sections, we explain how our algorithmic
result applies to these two formalisms.

\subsection{Mildly Context-Sensitive Language Recognition}
\label{section:tag}
\label{section:mildly-cs}

Linear context-free rewriting systems fall under the realm of mildly context-sensitive grammar formalisms. They subsume four important mildly context-sensitive formalisms
that were developed independently and later shown to be weakly equivalent by \citet{vw94}: tree-adjoining grammars \cite{joshi1997tree}, linear indexed grammars \cite{gazdar1988applicability},
head grammars \cite{pollard-84} and combinatory categorial grammars \cite{steedman-00}.
Weak equivalence here refers to the idea that any language generated by
a grammar in one of these formalisms can be also be generated by some grammar in any of the other formalisms among the four.
It can be verified that all of these formalisms are unbalanced, single-initial LCFRSs,
and as such, the algorithm in Figure~\ref{fig:parse-alg} applies to them.

\citet{rajasekaran98} showed that tree-adjoining grammars can be parsed with an asymptotic complexity of $O(M(n^2)) = O(n^{4.76})$. While he did not discuss that, the weak equivalence between
the four formalisms mentioned above implies that all of them can be parsed in time $O(M(n^2))$. Our algorithm generalizes this result. We now give the details.

Our starting point for this discussion is head grammars. Head grammars are a specific case of linear context-free rewriting systems, not just in the formal languages they define -- but also in the way
these grammars are described. They are described using concatenation production rules and wrapping production rules, which are directly transferable to LCFRS notation. Their fan-out is 2.
We focus in this discussion on ``binary head grammars,'' defined analogously to binary LCFRS -- the rank of all production rules has to be 2. The contact rank of binary head grammars is 2.
As such, our paper shows that the complexity of recognizing binary head grammar languages is $O(M(n^2)) = O(n^{4.76})$.

\citet{vw94} show that linear indexed grammars (LIGs) can actually be reduced to binary head grammars. Linear indexed grammars are extensions of CFGs, a linguistically-motivated restricted version of indexed
grammars, the latter of which were developed by \citet{aho1968indexed} for the goal of handling variable binding in programming languages. The main difference between LIGs and CFGs is that the nonterminals carry a ``stack,''
with a separate set of stack symbols. Production rules with LIGs copy the stack on the left-hand side to {\em one} of the nonterminal stacks in the righthand side,\footnote{General indexed grammars copy the stack
to multiple nonterminals on the right-hand side.} potentially pushing or popping one
symbol in the new copy of the stack. For our discussion, the main important detail about the reduction of LIGs to head grammars is that it preserves the rank of the production rules. As such, our paper
shows that binary LIGs can also be recognized in time $O(n^{4.76})$.

\citet{vw94} additionally address the issue of reducing combinatory categorial grammars to LIGs. The combinators they allow are function application and function composition. The key detail here is that their
reduction of CCG is to an LIG with rank 2, and as such, our algorithm applies to CCGs as well, which can be recognized in time $O(n^{4.76})$.

Finally, \citet{vw94} reduced tree-adjoining grammars to combinatory categorial grammars. The TAGs they tackle are in ``normal form,'' such that the auxiliary trees are binary (all TAGs can be reduced to normal form
TAGs). Such TAGs can be converted to weakly equivalent CCG (but not necessarily strongly equivalent), and as such, our algorithm applies to TAGs as well. As mentioned above, this finding supports the finding
of \citet{rajasekaran98}, who showed that TAG can be recognized in time $O(M(n^2))$.

For an earlier discussion connections between TAG parsing and Boolean matrix multiplication, see \citet{satta1994tree}.

\subsection{Synchronous Context-Free Grammars}
\label{section:scfg}

Synchronous Context-Free Grammars (SCFGs) are widely used in machine
translation to model the simultaneous derivation of translationally equivalent
strings in two natural languages, and are equivalent to the 
Syntax-Directed Translation Schemata of \citet{Aho:69a}.
SCFGs are a subclass of LCFRS where each nonterminal has
fan-out two: one span in one language and one span in the other.
Because the first span of the l.h.s.\ nonterminal always
contains spans from both r.h.s.\ nonterminals, SCFGs are always
single-initial.
Binary SCFGs, also known as Inversion Transduction Grammars (ITGs),
have no more than two nonterminals on the r.h.s.\ of a rule, and
are the most widely used model in syntax-based statistical machine translation.

Synchronous parsing with traditional tabular methods for ITG
is $O(n^6)$, as each of the three nonterminals in a rule has
fan-out of two.  ITGs, unfortunately, do not satisfy Condition~\ref{cond:abg}, and
therefore we have to use the algorithm in Figure~\ref{fig:parse-alg2}.
Still, just like with TAG, each rule combines two nonterminals of 
fan-out two using two combination points.  Thus, $d=2$, 
and we achieve a bound of $O(n^{2 \omega + 1})$ for ITG, which is $O(n^{5.76})$
using the current state of the art for matrix multiplication.

We achieve even greater gains for the case of multi-language synchronous
parsing.  Generalizing ITG to allow two nonterminals on the righthand
side of a rule in each of $k$ languages, we have an LCFRS with fan-out $k$. 
Traditional tabular parsing has an asymptotic complexity of $O(n^{3k})$, while
our algorithm has the complexity of $O(n^{\omega k +1})$.

Another interesting case of a synchronous formalism that our algorithm improves the best-well known result for is
that of binary synchronous TAGs \cite{shieber1990synchronous} -- i.e. a TAG in which all auxiliary trees are binary. This formalism
can be reduced to a binary LCFRS.
A tabular algorithm for such grammar has the asymptotic complexity of $O(n^{12})$. With our algorithm,
$d=4$ for this formalism, and as such its asymptotic complexity in that case is $O(n^{9.52})$.

\section{Discussion and Open Problems}

In this section, we discuss some extensions to our algorithm and open problems.

\subsection{Turning Recognition into Parsing}
\label{section:recognition-parsing}

The algorithm we presented focuses on recognition: given a string and a grammar, it can decide whether the string is in the language of the grammar
or not. From an application perspective, perhaps a more interesting algorithm is one that returns an actual derivation tree, if it identifies that
the string is in the language.

It is not difficult to adapt our algorithm to return such a parse, without changing the asymptotic complexity of $O(n^{\omega d + 1})$. Once the transitive closure
of $T$ is computed, we can backtrack to find such parse, starting with the start symbol in a cell spanning the whole string.
When we are in a specific cell, we check all possible combination points (there are $d$ of those) and nonterminals, and if we find such pairs of combination points
and nonterminals that are valid in the chart, then we backtrack to the corresponding cells. The asymptotic complexity of this post-processing step is $O(n^{d+1})$,
which is less than $O(n^{\omega d})$ ($\omega > 2$, $d > 1$).

This post-processing step corresponds to an algorithm that finds a parse tree, {\em given} a pre-calculated chart. If the chart was not already available when our algorithm
finishes, the asymptotic complexity of this step would correspond to the asymptotic complexity of a na\"ive tabular parsing algorithm.
It remains an open problem to adapt our algorithm to {\em probabilistic parsing}, for example -- finding the highest scoring parse given a probabilistic or a weighted LCFRS \cite{kallmeyer-maier:2010:PAPERS}.
See more details in \S\ref{section:wlp}.

\begin{figure}
\resizebox{\textwidth}{!}{
\begin{tabular}{p{5.0in}p{5.0in}}
\parpic{\begin{tikzpicture}
\draw(-1.5, 4.25)node{$D$};
\draw(-1.5, 3.25)node{$C$};
\draw(-1.5, 2.25)node{$B$};
\draw(-1.5, 1.25)node{$A$};
\draw(-1.5, 0.25)node{$S$};
\draw[fill=azure](0, 1)rectangle(1, 1.5);
\draw[fill=azure](1, 2)rectangle(2, 2.5);
\draw[fill=azure](2, 3)rectangle(3, 3.5);
\draw[fill=azure](3, 4)rectangle(4, 4.5);
\draw[fill=azure](5, 2)rectangle(6, 2.5);
\draw[fill=azure](6, 4)rectangle(7, 4.5);
\draw[fill=azure](7, 1)rectangle(8, 1.5);
\draw[fill=azure](8, 3)rectangle(9, 3.5);
\draw[fill=azure](-.2, 0.75)rectangle(9.2, 0.75); 
\draw[fill=azure](0, 0)rectangle(4, 0.5);
\draw[fill=azure](5, 0)rectangle(9, 0.5);
\end{tikzpicture} }
&
\parpic{\begin{tikzpicture}
\draw(-1.5, 4.25)node{$$};  
\draw(-1.5, 2.25)node{$B$};
\draw(-1.5, 1.25)node{$A$};
\draw(-1.5, 0.25)node{$(A,B)$};
\draw[fill=azure](0, 1)rectangle(1, 1.5);
\draw[fill=azure](1, 2)rectangle(2, 2.5);
\draw[fill=azure](5, 2)rectangle(6, 2.5);
\draw[fill=azure](7, 1)rectangle(8, 1.5);
\draw[fill=azure](-.2, 0.75)rectangle(9.2, 0.75); 
\draw[fill=azure](0, 0)rectangle(2, 0.5);
\draw[fill=azure](5, 0)rectangle(6, 0.5);
\draw[fill=azure](7, 0)rectangle(8, 0.5);
\end{tikzpicture} }
\\[2.5in]
\parpic{\begin{tikzpicture}
\draw(-1.5, 2.25)node{$C$};
\draw(-1.5, 1.25)node{$(A,B)$};
\draw(-1.5, 0.25)node{$(A,B,C)$};
\draw[fill=azure](2, 2)rectangle(3, 2.5); 
\draw[fill=azure](0, 1)rectangle(2, 1.5); 
\draw[fill=azure](8, 2)rectangle(9, 2.5); 
\draw[fill=azure](5, 1)rectangle(6, 1.5); 
\draw[fill=azure](7, 1)rectangle(8, 1.5); 
\draw[fill=azure](-.2, 0.75)rectangle(9.2, 0.75); 
\draw[fill=azure](0, 0)rectangle(3, 0.5);
\draw[fill=azure](5, 0)rectangle(6, 0.5);
\draw[fill=azure](7, 0)rectangle(9, 0.5);
\end{tikzpicture} }
&
\parpic{\begin{tikzpicture}
\draw(-1.5, 2.25)node{$D$};
\draw(-1.5, 1.25)node{$(A,B,C)$};
\draw(-1.5, 0.25)node{$S$};
\draw[fill=azure](3, 2)rectangle(4, 2.5); 
\draw[fill=azure](0, 1)rectangle(3, 1.5); 
\draw[fill=azure](6, 2)rectangle(7, 2.5); 
\draw[fill=azure](5, 1)rectangle(6, 1.5); 
\draw[fill=azure](7, 1)rectangle(9, 1.5); 
\draw[fill=azure](-.2, 0.75)rectangle(9.2, 0.75); 
\draw[fill=azure](0, 0)rectangle(4, 0.5);
\draw[fill=azure](5, 0)rectangle(9, 0.5);
\end{tikzpicture} }
\\[1.5in]
\end{tabular}}
\caption{Upper left: Combination of spans for 
SCFG rule $[ S \rightarrow A\, B\, C\, D,\,\,\, B\, D\, A\, C ]$.
Upper right and bottom row: three steps in
parsing binarized rule.}\label{fig:abcd}
\end{figure}

\subsection{General Recognition for Synchronous Parsing}

Similarly to LCFRS, the rank of an SCFG is the maximal number
of nonterminals that appear in the right-hand side of a rule.
Any SCFG can be binarized into 
an LCFRS grammar. However, when the SCFG rank is arbitrary, this means
that the fan-out of the LCFRS grammar can be larger than 2. This happens
because binarization creates intermediate nonterminals that
span several substrings, denoting binarization steps of the rule. These
substrings are eventually combined into two spans, to yield the language
of the SCFG grammar \cite{HuangZhangGildeaKnight-cl09}.


Our algorithm does not always improve the asymptotic complexity
of SCFG parsing over tabular methods. For example, Figure~\ref{fig:abcd} shows the
combination of spans for  the rule $[ S \rightarrow A\, B\, C\, D ,  B\, D\, A\, C ]$, along with 
a binarization into three simpler LCFRS rules. A na\"{i}ve tabular algorithm
for this rule would have the asymptotic complexity of $O(n^{10})$,
but the binarization shown in Figure~\ref{fig:abcd} reduces this to $O(n^8)$.
Our algorithm
gives a complexity of $O(n^{9.52})$, as the second step in the binarization 
shown consists of a rule with $d=4$.


\subsection{Generalization to Weighted Logic Programs}
\label{section:wlp}

Weighted logic programs (WLPs) are declarative programs, in the form of Horn clauses similar to those that Prolog uses, that can be used
to formulate parsing algorithms such as CKY and other types of dynamic programming algorithms or NLP inference algorithms \cite{eisner-05b,cohen-11a}.

For a given Horn clause, WLPs also require a ``join'' operation that sums (in some semiring) over a set of possible values in the free variables
in the Horn clauses. With CKY, for example, this sum will be performed on the mid-point concatenating two spans. This join operation is also
the type of operation we address in this paper (for LCFRS) in order to improve their asymptotic complexity.

It remains an open question to see whether we can generalize our algorithm to arbitrary weighted logic programs. In order to create an algorithm
that takes as input a weighted logic program (and a set of axioms) and ``recognizes'' whether the goal is achievable, we would need to have a generic
way of specifying the set $N$, which was specialized to LCFRS in this case. Not only that, we would have to specify $N$ in such a way that the asymptotic
complexity of the WLP would improve over a simple dynamic programming algorithm (or a memoization technique).

In addition, in this paper we focus on the problem of recognition and parsing for unweighted grammars.
\citet{benedi2007fast} showed how to generalize Valiant's algorithm in order to compute inside probabilities for a PCFG and a string.
Even if we were able to generalize our addressing scheme to WLPs, it remains an open question to see whether we can go beyond recognition (or
unweighted parsing).

\subsection{Rytter's Algorithm}

\citet{rytter1995context} gives an algorithm for CFG parsing with the same 
time complexity as Valiant's, but a somewhat simpler divide-and-conquer
strategy.  Rytter's algorithm works by first recursively finding 
all chart items entirely within the first half of the string 
and entirely within the second half of the string.  The combination 
step uses a shortest path computation to identify the 
sequence of chart items along a spine of the final
parse tree, where the spine extends from the root of the tree to the terminal in
position $n/2$.  Rytter's algorithm relies on the fact that 
this spine, consisting of chart items that cross the midpoint of the string,
forms a single path from the root to one leaf of the derivation tree.
This property does not hold for general LCFRS, since two siblings
in the derivation tree may both correspond to multiple 
spans in the string, each containing material on both sides of the 
string midpoint.  For this reason, Rytter's algorithm does not
appear to generalize easily to LCFRS.

\subsection{Relation to Multiple Context-Free Grammars}
\label{section:mcfg}

\citet{nakanishi1998efficient} developed a matrix multiplication parsing algorithm for multiple context-free grammars (MCFGs). When these grammars are given in a binary
form, they can be reduced to binary LCFRS. Similarly, binary LCFRS can be reduced to binary MCFGs. The algorithm that Nakanishi et al. develop is simpler
than ours, and does not directly tackle the problem of transitive closure for LCFRS.
More specifically, Nakanishi et al. multiply a seed matrix such as our $T$ by itself in several steps, and then follow up with a copying operation
between equivalent cells. They repeat this $n$ times, where $n$ is the sentence length. As such, the asymptotic complexity of their algorithm is identical
for both balanced and unbalanced grammars, a distinction they do not make.

The complexity analysis of Nakanishi et al. is different than ours, but in certain cases, yields identical results. For example, if $\varphi(a) = f$ for all $a \in \nts$,
and the grammar is balanced, then both our algorithm and their algorithm give a complexity of $O(n^{\omega f + 1})$. If the grammar is unbalanced, then our algorithm
gives a complexity of $O(n^{\omega f})$, while the asymptotic complexity of their algorithm remains $O(n^{\omega f + 1})$. As such, Nakanishi et al.'s algorithm does
not generalize Valiant's algorithm -- its asymptotic complexity for context-free grammars is $O(n^{\omega + 1})$ and not $O(n^{\omega})$.

Nakanishi et al. pose in their paper an open problem, which loosely can be reworded as the problem of finding an algorithm that computes the transitive closure of $T$
without the extra $O(n)$ factor that their algorithm incurs. In our paper, we provide a solution to this open problem for the case of single-initial,
unbalanced grammars. The
core of the solution lies in the matrix multiplication copying mechanism described in \S\ref{section:mul}.

\subsection{Optimal Binarization Strategies}
\label{section:binarization-strategy}

The two main grammar parameters that affect the asymptotic complexity of parsing with LCFRS (in its general form) are the fan-out of the nonterminals and
the rank of the rules. With tabular parsing, we can actually refer to the parsing complexity of a {\em specific rule in the grammar}. Its complexity is
$O(n^{p})$, where the parsing complexity $p$ is
the total fan-out of all nonterminals in the rule.
For binary rules of the form $A \rightarrow B\, C$,
$p= \varphi(A) + \varphi(B) + \varphi(C)$.

To optimize the tabular algorithm time complexity of parsing with
a binary LCFRS, equivalent to another non-binary LCFRS, we would
want to minimize the time complexity it takes to parse each
rule. As such, our goal is to minimize $\varphi(A) + \varphi(B) +
\varphi(C)$ in the resulting binary grammar.  \citet{gildea-cl11}
has shown that this metric corresponds to the tree width of a
dependency graph which is constructed from the grammar.   It 
is not known whether finding the optimal binarization of 
an LCFRS is an NP-complete problem, but \citet{gildea-cl11}
shows that a polynomial time algorithm would imply improved
approximation algorithms for the treewidth of general graphs.

In general, the optimal binarization for tabular parsing may not
by the same as the optimal binarization for parsing with our
algorithm based on matrix multiplication.
In order to optimize the
complexity of our algorithm, we want to minimize $d$, which is
the maximum over all rules $A \rightarrow B\, C$ of
\begin{equation}
d(A\rightarrow B\, C) = \max \{ \varphi(A) + \varphi(B) -
\varphi(C), \varphi(A) - \varphi(B) + \varphi(C), -\varphi(A) +
\varphi(B) + \varphi(C) \}.
\end{equation}

For a fixed binarized grammar, $d$ is always less than $p$, the tabular
parsing complexity, and, hence, the optimal $d^*$ over binarizations of an LCFRS
is always less than the optimal $p^*$ for tabular parsing.
However, whether any savings can be achieved with our algorithm 
depends on whether $\omega d^* < p^*$, or $\omega d^* + 1 < p^*$
in the case of balanced grammars.
Our criterion does not seem to correspond closely to a
well-studied graph-theoretic concept such a treewidth, and 
it remains an open problem to find
an efficient algorithm that minimizes this definition of parsing complexity.

It is worth noting that $d(A \rightarrow B\, C) \ge \frac{1}{3} \left( \varphi(A) + \varphi(B) + \varphi(C) \right)$. As such, this gives a lower bound on the time complexity 
of our algorithm relative to tabular parsing using the same binarized
grammar. If $O(n^{t_1})$ is the asymptotic complexity of our algorithm, and $O(n^{t_2})$ is the asymptotic complexity of a tabular
algorithm, then $\displaystyle\frac{t_1}{t_2} \ge \frac{\omega}{3} > 0.79$.


\section{Conclusion}

We described a parsing algorithm for binary linear context-free rewriting systems that has the asymptotic complexity of $O(n^{\omega d + 1})$ where $\omega < 2.38$,
$d$ is the ``contact rank'' of the grammar (the maximal number of combination points in the rules in the grammar
in single-initial form) and $n$ is the string length. Our algorithm
has the asymptotic complexity of $O(n^{\omega d})$ for a subset of binary LCFRS which are unbalanced. Our result generalizes the algorithm of \citet{valiant75}, and also
reinforces existing results about mildly context-sensitive parsing for tree-adjoining grammars \cite{rajasekaran98}. Our result also implies that inversion transduction
grammars can be parsed in time $O(n^{2\omega + 1})$ and that synchronous parsing with $k$ languages has the asymptotic complexity of $O(n^{\omega k +1})$ where $k$ is the number
of languages.

\appendix
\setcounter{section}{0}
\renewcommand{\theequation}{\Alph{section}.\arabic{equation}}
\renewcommand{\thesection}{\Alph{section}}
\renewcommand{\thesection}{\Alph{section}}
\renewcommand{\appendixsection}[1]{%
   \setcounter{table}{0}
   \setcounter{figure}{0}
   \setcounter{equation}{0}
  \section{#1}%
}

\appendixsection{Notation}\label{notation}
Table~\ref{table:notation} gives a table of notation for symbols used throughout this paper.

\begin{table}
\begin{tabular}{|l|l|l|}
\hline
Symbol & Description & 1st mention \\
\hline
$M(n)$ & The complexity of Boolean $n \times n$ matrix multiplication & \S\ref{section:introduction} \\
$\omega$ & Best well-known complexity for $M(n)$, $M(n) = O(n^\omega)$ & \S\ref{section:introduction} \\
$[n]$ & Set of integers $\{ 1,\dots,n\}$ & \S\ref{section:notation} \\
$[n]_0$ & $[n] \cup \{ 0 \}$ & \S\ref{section:notation} \\
$\nts$ & Nonterminals of the LCFRS & \S\ref{section:notation} \\
$\vocab$ & Terminal symbols of the LCFRS & \S\ref{section:notation} \\
$\vars$ & Variables that denote spans in grammar & \S\ref{section:notation} \\
$\rulessym$ & Rules in the LCFRS & \S\ref{section:notation} \\
$A$,$B$,$C$ & Nonterminals & \S\ref{section:notation} \\
$f$ & Maximal fan-out of the LCFRS & Eq.~\refeq{eq:f} \\
$\varphi(A)$ & Fan-out of nonterminal $A$ & \S\ref{section:notation} \\
$y$ & Denoting a variable in $\vars$ (potentially subscripted) & \S\ref{section:notation} \\
$T$ & Seed matrix & \S\ref{section:sketch} \\
$N$, $N(d)$ & Set of indices for addresses in the matrix & Eq.~\refeq{eq:nd} \\
$i$, $j$ & Indices for cells in $T$. $i,j \in N$ & \S\ref{section:matrix-structure}\\
$d$ & Grammar contact rank & \S\ref{section:matrix-structure} \\
$M$ & $T_{ij}$ is a subset of $M$ & \S\ref{section:matrix-structure} \\
$\frow$,$\trow$,$\urow$ & Copying/marking symbols for rows & \S\ref{section:matrix-structure}\\
$\tcol$,$\fcol$,$\ucol$ & Copying/marking symbols for columns & \S\ref{section:matrix-structure} \\
$n$ & Length of sentence to be parsed & \S\ref{section:introduction} \\
$<$ & Total order between the set of indices of $T$ & \S\ref{section:matrix-structure}\\
$m(i,j)$ & Merged sorted sequence of $i \cup j$, divided into pairs & \S\ref{section:matrix-structure} \\
$\remove(v,x)$ & Removal of $x$ from a sequence $v$ & Figure~\ref{fig:otimes} \\
$\insertt(v,x)$ & Insertion of $x$ in a sequence $v$ & \S\ref{section:transitive-closure} \\
$\Pi$ & Copying operator & \S\ref{section:transitive-closure} \\
\hline
\end{tabular}

\caption{\label{table:notation} Table of notation symbols used in this paper.}
\end{table}

\section*{Acknowledgments}
The authors thank the anonymous reviewers for their comments and
Adam Lopez and Giorgio Satta for useful conversations.
This work was supported by NSF grant IIS-1446996 and by EPSRC grant EP/L02411X/1.

\twocolumn
\bibliographystyle{fullname}
\bibliography{all}

\begin{thebibliography}{}

\bibitem[\protect\citename{Abboud, Backurs, and Williams}2015]{abboud2015if}
Abboud, Amir, Arturs Backurs, and Virginia~Vassilevska Williams.
\newblock 2015.
\newblock If the current clique algorithms are optimal, so is {Valiant's}
  parser.
\newblock {\em arXiv preprint arXiv:1504.01431}.

\bibitem[\protect\citename{Aho}1968]{aho1968indexed}
Aho, Alfred~V.
\newblock 1968.
\newblock Indexed grammars -- an extension of context-free grammars.
\newblock {\em Journal of the ACM (JACM)}, 15(4):647--671.

\bibitem[\protect\citename{Aho and Ullman}1969]{Aho:69a}
Aho, Alfred~V. and Jeffery~D. Ullman.
\newblock 1969.
\newblock Syntax directed translations and the pushdown assembler.
\newblock {\em Jounral of Computer and System Sciences}, 3:37--56.

\bibitem[\protect\citename{Bened{\'\i} and S{\'a}nchez}2007]{benedi2007fast}
Bened{\'\i}, Jos{\'e}-Miguel and Joan-Andreu S{\'a}nchez.
\newblock 2007.
\newblock Fast stochastic context-free parsing: A stochastic version of the
  {Valiant} algorithm.
\newblock In {\em Pattern Recognition and Image Analysis}. Springer, pages
  80--88.

\bibitem[\protect\citename{Cocke and Schwartz}1970]{cocke-70}
Cocke, John and Jacob~T. Schwartz.
\newblock 1970.
\newblock Programming languages and their compilers: Preliminary notes.
\newblock Technical report, Courant Institute of Mathematical Sciences, New
  York University.

\bibitem[\protect\citename{Cohen, Satta, and Collins}2013]{cohen-13a}
Cohen, Shay~B., Giorgio Satta, and Michael Collins.
\newblock 2013.
\newblock Approximate {PCFG} parsing using tensor decomposition.
\newblock In {\em Proceedings of the 2013 Meeting of the North American chapter
  of the Association for Computational Linguistics (NAACL-13)}, pages 487--496.

\bibitem[\protect\citename{Cohen, Simmons, and Smith}2011]{cohen-11a}
Cohen, Shay~B., Robert~J. Simmons, and Noah~A. Smith.
\newblock 2011.
\newblock Products of weighted logic programs.
\newblock {\em Theory and Practice of Logic Programming}, 11(2--3):263--296.

\bibitem[\protect\citename{Coppersmith and Winograd}1987]{CoppersmithW87}
Coppersmith, D. and S.~Winograd.
\newblock 1987.
\newblock Matrix multiplication via arithmetic progressions.
\newblock In {\em Proceedings of the 19-th annual ACM conference on {T}heory of
  computing}, pages 1--6.

\bibitem[\protect\citename{Dunlop, Bodenstab, and
  Roark}2010]{dunlop2010reducing}
Dunlop, Aaron, Nathan Bodenstab, and Brian Roark.
\newblock 2010.
\newblock Reducing the grammar constant: an analysis of {CYK} parsing
  efficiency.
\newblock Technical report, Technical report CSLU-2010-02, OHSU.

\bibitem[\protect\citename{Earley}1970]{earley1970efficient}
Earley, Jay.
\newblock 1970.
\newblock An efficient context-free parsing algorithm.
\newblock {\em Communications of the ACM}, 13(2):94--102.

\bibitem[\protect\citename{Eisner, Goldlust, and Smith}2005]{eisner-05b}
Eisner, Jason, Eric Goldlust, and Noah~A. Smith.
\newblock 2005.
\newblock Compiling {Comp Ling}: Practical weighted dynamic programming and the
  {Dyna} language.
\newblock In {\em Proceedings of {HLT-EMNLP}}, pages 281--290.

\bibitem[\protect\citename{Eisner and Satta}1999]{eisner1999efficient}
Eisner, Jason and Giorgio Satta.
\newblock 1999.
\newblock Efficient parsing for bilexical context-free grammars and head
  automaton grammars.
\newblock In {\em Proceedings of the 37th annual meeting of the Association for
  Computational Linguistics on Computational Linguistics}, pages 457--464.
  Association for Computational Linguistics.

\bibitem[\protect\citename{Gazdar}1988]{gazdar1988applicability}
Gazdar, Gerald.
\newblock 1988.
\newblock {\em Applicability of indexed grammars to natural languages}.
\newblock Springer.

\bibitem[\protect\citename{Gildea}2011]{gildea-cl11}
Gildea, Daniel.
\newblock 2011.
\newblock Grammar factorization by tree decomposition.
\newblock {\em Computational Linguistics}, 37(1):231--248.

\bibitem[\protect\citename{Huang \bgroup et al.\egroup
  }2009]{HuangZhangGildeaKnight-cl09}
Huang, Liang, Hao Zhang, Daniel Gildea, and Kevin Knight.
\newblock 2009.
\newblock Binarization of synchronous context-free grammars.
\newblock {\em Computational Linguistics}, 35(4):559--595.

\bibitem[\protect\citename{Joshi and Schabes}1997]{joshi1997tree}
Joshi, Aravind~K and Yves Schabes.
\newblock 1997.
\newblock Tree-adjoining grammars.
\newblock In {\em Handbook of formal languages}. Springer, pages 69--123.

\bibitem[\protect\citename{Kallmeyer}2010]{kallmeyer-10}
Kallmeyer, Laura.
\newblock 2010.
\newblock {\em Parsing Beyond Context-Free Grammars}.
\newblock Cognitive Technologies. Springer.

\bibitem[\protect\citename{Kallmeyer and
  Maier}2010]{kallmeyer-maier:2010:PAPERS}
Kallmeyer, Laura and Wolfgang Maier.
\newblock 2010.
\newblock Data-driven parsing with probabilistic linear context-free rewriting
  systems.
\newblock In {\em Proceedings of the 23rd International Conference on
  Computational Linguistics ({COLING} 2010)}, pages 537--545.

\bibitem[\protect\citename{Kasami}1965]{kasami-65}
Kasami, Tadao.
\newblock 1965.
\newblock An efficient recognition and syntax-analysis algorithm for
  context-free languages.
\newblock Technical Report AFCRL-65-758, Air Force Cambridge Research Lab.

\bibitem[\protect\citename{Le~Gall}2014]{LeGall:2014}
Le~Gall, Fran\c{c}ois.
\newblock 2014.
\newblock Powers of tensors and fast matrix multiplication.
\newblock In {\em Proceedings of the 39th International Symposium on Symbolic
  and Algebraic Computation}, ISSAC '14, pages 296--303, New York, NY, USA.
  ACM.

\bibitem[\protect\citename{Nakanishi \bgroup et al.\egroup
  }1998]{nakanishi1998efficient}
Nakanishi, Ryuichi, Keita Takada, Hideki Nii, and Hiroyuki Seki.
\newblock 1998.
\newblock Efficient recognition algorithms for parallel multiple context-free
  languages and for multiple context-free languages.
\newblock {\em IEICE TRANSACTIONS on Information and Systems},
  81(11):1148--1161.

\bibitem[\protect\citename{Pollard}1984]{pollard-84}
Pollard, Carl~J.
\newblock 1984.
\newblock {\em Generalized Phrase Structure Grammars, Head Grammars and Natural
  Languages}.
\newblock {Ph.D.} thesis, Stanford University.

\bibitem[\protect\citename{Rajasekaran and Yooseph}1998]{rajasekaran98}
Rajasekaran, Sanguthevar and Shibu Yooseph.
\newblock 1998.
\newblock {TAL} parsing in {$O(M(n^2))$} time.
\newblock {\em Journal of Computer and System Sciences}, 56:83--89.

\bibitem[\protect\citename{Raz}2002]{raz2002complexity}
Raz, Ran.
\newblock 2002.
\newblock On the complexity of matrix product.
\newblock In {\em Proceedings of the thiry-fourth annual ACM symposium on
  Theory of computing}, pages 144--151. ACM.

\bibitem[\protect\citename{Rytter}1995]{rytter1995context}
Rytter, Wojciech.
\newblock 1995.
\newblock Context-free recognition via shortest paths computation: a version of
  {Valiant's} algorithm.
\newblock {\em Theoretical Computer Science}, 143(2):343--352.

\bibitem[\protect\citename{Satta}1992]{satta1992recognition}
Satta, Giorgio.
\newblock 1992.
\newblock Recognition of linear context-free rewriting systems.
\newblock In {\em Proceedings of the 30th annual meeting on Association for
  Computational Linguistics}, pages 89--95. Association for Computational
  Linguistics.

\bibitem[\protect\citename{Satta}1994]{satta1994tree}
Satta, Giorgio.
\newblock 1994.
\newblock Tree-adjoining grammar parsing and boolean matrix multiplication.
\newblock {\em Computational Linguistics}, 20(2):173--191.

\bibitem[\protect\citename{Shieber}1985]{Shieber85}
Shieber, S.~M.
\newblock 1985.
\newblock Evidence against the context-freeness of natural language.
\newblock In {\em Linguistics and Philosophy}, volume~8. D. Reidel Publishing
  Company, pages 333--343.

\bibitem[\protect\citename{Shieber and Schabes}1990]{shieber1990synchronous}
Shieber, Stuart~M and Yves Schabes.
\newblock 1990.
\newblock Synchronous tree-adjoining grammars.
\newblock In {\em Proceedings of the 13th conference on Computational
  linguistics-Volume 3}, pages 253--258. Association for Computational
  Linguistics.

\bibitem[\protect\citename{Steedman}2000]{steedman-00}
Steedman, Mark.
\newblock 2000.
\newblock {\em The Syntactic Process}.
\newblock Language, speech, and communication. MIT Press, Cambridge (Mass.),
  London.

\bibitem[\protect\citename{Strassen}1969]{Strassen69}
Strassen, V.
\newblock 1969.
\newblock Gaussian elimination is not optimal.
\newblock {\em Numerische Mathematik}, 14(3):354--356.

\bibitem[\protect\citename{Valiant}1975]{valiant75}
Valiant, Leslie~G.
\newblock 1975.
\newblock General context-free recognition in less than cubic time.
\newblock {\em Journal of Computer and System Sciences}, 10:308--315.

\bibitem[\protect\citename{Vijay-Shanker and Weir}1994]{vw94}
Vijay-Shanker, K. and David Weir.
\newblock 1994.
\newblock The equivalence of four extensions of context-free grammars.
\newblock {\em Mathematical Systems Theory}, 27:511--546.

\bibitem[\protect\citename{Wu}1997]{wu1997stochastic}
Wu, Dekai.
\newblock 1997.
\newblock Stochastic inversion transduction grammars and bilingual parsing of
  parallel corpora.
\newblock {\em Computational Linguistics}, 23(3):377--403.

\bibitem[\protect\citename{Younger}1967]{younger-67}
Younger, Daniel~H.
\newblock 1967.
\newblock Recognition and parsing of context-free languages in time $n^3$.
\newblock {\em Information and Control}, 10(2):189--208.

\end{thebibliography}

\end{document}